\documentclass[10pt,twocolumn,letterpaper]{article}

\usepackage{iccv}
\usepackage{times}
\usepackage{epsfig}
\usepackage{graphicx}
\usepackage{amsmath}
\usepackage{amssymb}

\usepackage{multirow}
\usepackage{pifont}
\usepackage{booktabs}
\usepackage{graphics}
\usepackage{threeparttable}
\usepackage{color}
\usepackage[normalem]{ulem}
\usepackage{float}
\usepackage{amsfonts}
\usepackage{bm}
\usepackage{subfig}
\usepackage{enumitem}
\usepackage[numbers]{natbib}
\usepackage{array}
\usepackage[table]{xcolor}
\usepackage{colortbl}
\usepackage[export]{adjustbox}
\usepackage{gensymb}
\usepackage{soul}
\usepackage{cancel}
\usepackage{algorithm}

\usepackage[pagebackref=true,breaklinks=true,letterpaper=true,colorlinks,bookmarks=false]{hyperref}

\iccvfinalcopy 

\newcolumntype{I}{!{\vrule width 1pt}}

\DeclareMathOperator*{\argmax}{argmax}

\newcommand{\makesupptitle}[1]{
	\twocolumn[
	\begin{center}
		{\Large \bf #1 \par}
		{
		\large
		\lineskip .5em
		\par
		}
		\vskip .5em
		\vspace*{12pt}
	\end{center}
	]
}

\newcommand{\red}[1]{\textcolor{red}{#1}}
\newcommand{\blue}[1]{\textcolor{blue}{#1}}

\makeatletter
\newcommand{\thickhline}{%
    \noalign {\ifnum 0=`}\fi \hrule height 1pt
    \futurelet \reserved@a \@xhline
}
\makeatother

\newcommand{\sssection}[1]{\noindent\textbf{#1}}
\newcommand{\pub}[1]{\color{gray}{\tiny{[{#1}]}}}

\usepackage{caption}
\usepackage[utf8]{inputenc}

\definecolor{bblue}{RGB}{0,30,95}
\definecolor{rred}{RGB}{190,0,0}
\definecolor{mygray}{gray}{.9}
\definecolor{ggray}{RGB}{127,127,127}
\definecolor{sblue}{RGB}{0,173,206}
\definecolor{ppink}{RGB}{240,46,142}

\usepackage{cleveref}
\crefname{section}{§}{§§}
\Crefname{section}{§}{§§}

\def\iccvPaperID{2447} 

\ificcvfinal\fi

\begin{document}

\title{\textsc{Dreamwalker}: Mental Planning for Continuous Vision-Language Navigation}

\author{
Hanqing Wang\textsuperscript{\textnormal{1, 2}}\qquad
Wei Liang\textsuperscript{\textnormal{1,4}$^*$}\qquad
Luc Van Gool\textsuperscript{\textnormal{2}}\qquad
Wenguan Wang\textsuperscript{\textnormal{3}\thanks{Corresponding authors.}}\\
\small \textsuperscript{\textnormal{1}}Beijing Institute of Technology \ \ \small \textsuperscript{\textnormal{2}}Computer Vision Lab, ETH Zurich \ \ \small \textsuperscript{\textnormal{3}}ReLER, CCAI, Zhejiang University\\
\small \textsuperscript{\textnormal{4}}Yangtze Delta Region Academy of Beijing Institute of Technology, Jiaxing\\
\small \url{https://github.com/hanqingwangai/Dreamwalker}
}

\maketitle
\ificcvfinal\thispagestyle{empty}\fi

\begin{abstract}
   VLN-CE is a recently released embodied task, where~AI agents need to navigate a freely traversable environment to reach a distant target location, given language instructions. It poses great challenges due to the huge space of possible strategies. Driven by the belief that the ability to anticipate the consequences of future actions is crucial for the emergence of intelligent and interpretable planning behavior, we propose \textsc{Dreamwalker} --- a \textbf{{world model}} based VLN-CE  agent. The world model is built to summarize~the visual,~to- pological, and dynamic properties of the complicated conti- nuous environment$_{\!}$ into$_{\!}$ a$_{\!}$ discrete,$_{\!}$ structured, and$_{\!}$ compact$_{\!}$ representation.$_{\!}$ \textsc{Dreamwalker}$_{\!}$~can$_{\!}$~simulate and evaluate possible plans entirely in such internal abstract world, before executing costly actions. As opposed to existing~model-free$_{\!}$ VLN-CE agents$_{\!}$ simply$_{\!}$ making$_{\!}$ greedy$_{\!}$ decisions$_{\!}$ in$_{\!}$ the$_{\!}$ real$_{\!}$ world, which easily results in shortsighted behaviors, \textsc{Dreamwalker} is able to make strategic planning through large amounts of ``mental experiments.'' Moreover, the ima- gined future scenarios reflect our agent's intention, making$_{\!}$ its$_{\!}$ decision-making$_{\!}$ process$_{\!}$ more$_{\!}$ transparent.$_{\!}$ Extensive$_{\!}$~ex-
   periments and ablation studies on VLN-CE dataset confirm the effectiveness of the proposed approach and outline fruitful directions for future work. 
\end{abstract}

\section{Introduction}

For decades, the AI community has strived to develop intelligent robots that can understand human instructions and carry them out. As a small step towards this long-held goal, vision-language navigation (VLN)~\cite{anderson2018vision} --- the task of entailing autonomous agents to navigate in \textit{never-before-seen} 3D environments$_{\!}$ with$_{\!}$ language$_{\!}$ instructions$_{\!}$ ---$_{\!}$ gained$_{\!}$ growing  attention. In the standard VLN setting, agent's movement is$_{\!}$ constrained$_{\!}$ to$_{\!}$ a$_{\!}$ small$_{\!}$ set$_{\!}$ of$_{\!}$ pre-defined$_{\!}$ sparse$_{\!}$ locations. As pointed out by~\cite{krantz2020navgraph}, such over-simplified, discrete task setup involves many unrealistic assumptions such as known topology,$_{\!}$ perfect$_{\!}$ localization,$_{\!}$ and$_{\!}$ deterministic$_{\!}$ transition.

To better reflect the challenges of real world navigation, Krantz$_{\!}$ \etal$_{\!\!\!}$~\cite{krantz2020navgraph}$_{\!}$ update$_{\!}$ the$_{\!}$ discrete$_{\!}$ VLN$_{\!}$ to$_{\!}$ a$_{\!}$ continuous$_{\!}$ ver- sion$_{\!}$ --$_{\!\!}$ VLN-CE$_{\!}$ (VLN$_{\!}$ in$_{\!}$ continuous$_{\!}$ environments),$_{\!}$ where  the agent is free to traverse any unobstructed location with low-level actions.$_{\!}$ VLN-CE~proved much more challenging than its discrete counterpart:$_{\!}$ the
performance$_{\!}$ gap$_{\!}$ between$_{\!}$ the$_{\!}$ state-of-the-arts$_{\!}$ in$_{\!}$ the$_{\!}$ two$_{\!}$ settings is more than 20\%, in terms of episode success rate. The main challenge posed by VLN-CE~lies in
the demand of strategic planning in conti- nuous environments with low-level$_{\!}$ actions.

\begin{figure}
	\vspace{-12pt}
	\begin{center}
		\includegraphics[width=0.9\linewidth]{fig/fig1}
	\end{center}
	\vspace{-16pt}
	\captionsetup{font=small}
	\caption{\small In partially observable, continuous VLN environments, \textsc{Dreamwalker}$_{\!}$ maps$_{\!}$ its$_{\!}$ surrounding$_{\!}$ into$_{\!}$ a$_{\!}$ discrete$_{\!}$ and$_{\!}$ structured$_{\!}$ abstraction.$_{\!}$ In this internal world, it is able to conduct mental plan- ning$_{\!}$ (\protect\includegraphics[scale=0.07,valign=c]{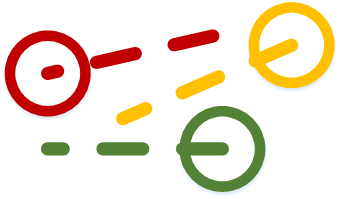})$_{\!}$ by$_{\!}$ imagining$_{\!}$ future$_{\!}$ scenarios,$_{\!}$ before$_{\!}$ taking$_{\!}$ real$_{\!}$ action.}
	\label{fig:teaser}
	\vspace{-13pt}
\end{figure}

As a direct response, we developed a \textit{world model} based VLN-CE$_{\!}$ agent,$_{\!}$ called$_{\!}$ \textsc{Dreamwalker}.$_{\!}$ Previous$_{\!}$ studies$_{\!}$~in cognitive science~\cite{forrester1971counterintuitive,johnson1983mental,johnson2010mental} suggest that humans build a mental model of the local surrounding, based on our limited senses. This internal world model summarizes our knowle- dge about the environment and serves as the basis for many high-level meta-skills, \eg, reasoning, planning, decision-making, and interpretation. The world model theory is one source of the idea of model-based Reinforcement Learning (RL)$_{\!}$~\cite{silver2016mastering}$_{\!}$ and$_{\!}$ promotes$_{\!}$ many$_{\!}$ recent$_{\!}$ advances$_{\!}$ in$_{\!}$ robot$_{\!}$ con-  trol~\cite{rybkin2018learning,piergiovanni2019learning,nair2020goal,yen2020experience}. Keeping this grand idea in head, we let \textsc{Dreamwalker}$_{\!}$ explicitly$_{\!}$ abstract$_{\!}$ crucial$_{\!}$ characteristics$_{\!}$~of its$_{\!}$~continuous$_{\!}$ surrounding$_{\!}$ environment$_{\!}$ to$_{\!}$ a$_{\!}$ \textit{discrete},$_{\!}$ \textit{struc-} \textit{tured}$_{\!}$ representation$_{\!}$ (Fig.$_{\!\!}$~\ref{fig:teaser}).$_{\!}$ This$_{\!}$ allows$_{\!}$ \textsc{Dreamwalker}$_{\!}$ to ``imagine'' a lot of future possible navigation plans and eva- luate the corresponding consequences entirely in the mind, before taking actual low-level actions in the real world.~In this$_{\!}$ way,$_{\!}$ \textsc{Dreamwalker}$_{\!}$ takes$_{\!}$ the$_{\!}$ challenge$_{\!}$ of$_{\!}$ VLN-CE head-on:$_{\!}$ mental$_{\!}$ planning$_{\!}$ with$_{\!}$ discrete$_{\!}$ world$_{\!}$ model$_{\!}$ enables efficient navigation behavior in continuous environments.

Technically,$_{\!}$ the$_{\!}$ world$_{\!}$ model$_{\!}$ is$_{\!}$ built$_{\!}$ upon$_{\!}$ agent's$_{\!}$ past experiences and can make predictions about the future. It~con- tains$_{\!}$ two$_{\!}$ parts:$_{\!}$ \textbf{i)}$_{\!}$ An$_{\!}$ \textit{environment$_{\!}$ graph}$_{\!}$ (EG)$_{\!}$ is$_{\!}$ constructed as$_{\!}$ a$_{\!}$ composition$_{\!}$ of$_{\!}$ selected$_{\!}$ or$_{\!}$ predicted$_{\!}$ waypoints$_{\!}$ and$_{\!}$ their typological relations. EG collects agent's temporary know- ledge$_{\!}$ about$_{\!}$ its$_{\!}$ surrounding.$_{\!}$ \textbf{ii)}$_{\!}$ A$_{\!}$ learnable$_{\!}$ \textit{scene$_{\!}$ synthesizer}
(SS)$_{\!}$ predicts$_{\!}$ future$_{\!}$ observations$_{\!}$ from$_{\!}$ a$_{\!}$ waypoint$_{\!}$ with$_{\!}$ mul-
 tiple$_{\!}$ steps. SS embeds agent's stable knowledge about envi-
 ronments,$_{\!}$ such$_{\!}$ as$_{\!}$ general$_{\!}$ room$_{\!}$ layout$_{\!}$ rules$_{\!}$ and$_{\!}$  transition dynamics,$_{\!}$ into$_{\!}$ its$_{\!}$ network$_{\!}$ parameters.$_{\!}$ Based on the$_{\!}$ world
 model,$_{\!}$ \textsc{Dreamwalker}$_{\!}$ synthesizes$_{\!}$ various$_{\!}$ future$_{\!}$ navigation trajectories,$_{\!}$ and$_{\!}$ assesses$_{\!}$ their$_{\!}$ progress$_{\!}$ towards the final target$_{\!}$ location.$_{\!}$ Then,$_{\!}$ the$_{\!}$ best$_{\!}$ mental$_{\!}$ plan$_{\!}$ is$_{\!}$ found$_{\!}$ by$_{\!}$ Monte Carlo$_{\!}$ Tree$_{\!}$ Search$_{\!}$~\cite{kocsis2006bandit} and executed in the continuous$_{\!}$ world with low-level$_{\!}$ actions.$_{\!}$ With$_{\!}$ the$_{\!}$ navigation$_{\!}$  proceeds,$_{\!}$ EG$_{\!}$ is$_{\!}$ further updated for making a new round of mental planning.

Notably,$_{\!}$ our$_{\!}$ \textsc{Dreamwalker}$_{\!}$ significantly$_{\!}$ distinguishes itself$_{\!}$ from$_{\!}$ prior$_{\!}$ VLN-CE$_{\!\!}$ solutions$_{\!}$~\cite{raychaudhuri2021language,hong2022bridging,krantz2021waypoint,krantz2022sim}$_{\!\!}$  in$_{\!}$ the following$_{\!}$ aspects:$_{\!}$ \textbf{i)}$_{\!}$ Recent$_{\!}$ advanced$_{\!}$ solutions$_{\!}$ are$_{\!}$ essentially model-free methods. While in principle a representation of the environment could be \textit{implicitly} learned through model-free RL, the reinforcement$_{\!}$ signal$_{\!}$ may$_{\!}$ be$_{\!}$ too$_{\!}$ weak~to$_{\!}$ quickly$_{\!}$ learn$_{\!}$ such$_{\!}$ a$_{\!}$ representation$_{\!}$ and$_{\!}$ how$_{\!}$ to$_{\!}$ make$_{\!}$ use$_{\!}$ of$_{\!}$ it. In contrast, our agent plans its actions within an explicit, and abstract model of the continuous environment.  \textbf{ii)} Existing agents navigate by greedily and reactively choosing between a small set of nearby waypoints, based on their hidden state which compresses past$_{\!}$ observations.$_{\!}$ They$_{\!}$ tend$_{\!}$ to be$_{\!}$ shortsighted, due to the absence of reliable strategies for capturing information for$_{\!}$ achieving$_{\!}$ the$_{\!}$ future$_{\!}$~\cite{hafner2019dream}.$_{\!}$ Yet \textsc{Dreamwalker} can use the world model to anticipate the impacts of possible actions and plan strategic behavior.$_{\!}$
\textbf{iii)}$_{\!}$ The$_{\!}$ future$_{\!}$ scenarios$_{\!}$ created$_{\!}$ by$_{\!}$ the$_{\!}$ world$_{\!}$ model$_{\!}$ explain$_{\!}$ the intention of  \textsc{Dreamwalker} in a way that human can understand, making its behaviors more interpretable \cite{chen2021interpretable,wang2021interpretable}.

Extensive experiments on VLN-CE dataset~\cite{krantz2020navgraph} confirm$_{\!}$ that$_{\!}$ our$_{\!}$ \textsc{Dreamwalker}$_{\!}$ gains$_{\!}$ promising$_{\!}$ performance$_{\!}$  with the appealing ability of real-time behavioral$_{\!}$ interpretation. This$_{\!}$ work$_{\!}$ is$_{\!}$ expected$_{\!}$ to$_{\!}$ foster$_{\!}$ future$_{\!}$~research$_{\!}$ in$_{\!}$ developing$_{\!}$ more$_{\!}$ strategic,$_{\!}$ robust,$_{\!}$ and$_{\!}$ interpretable VLN-CE agents.

\section{Related Work}
\sssection{VLN$_{\!}$ in$_{\!}$ Discrete$_{\!}$ Environments.$_{\!}$} The$_{\!}$ release$_{\!}$ of$_{\!}$ R2R$_{\!}$ data- set$_{\!}$~\cite{anderson2018vision}$_{\!}$ has$_{\!}$ stimulated$_{\!}$ the$_{\!}$ emergence$_{\!}$ of$_{\!}$ various$_{\!\!}$ VLN$_{\!}$ systems and$_{\!}$ datasets.$_{\!}$ In$_{\!}$ particular,$_{\!}$~early$_{\!}$ VLN$_{\!}$ agents$_{\!}$ are$_{\!}$ built$_{\!}$ upon recurrent neural network based sequence-to-sequence~mo- del$_{\!}$~\cite{anderson2018vision,fried2018speaker,tan2019learning,wang2018look,wang2019reinforced,huang2019transferable,ma2019self,zhu2019vision,ke2019tactical,ma2019regretful,wang2020active,wang2023active,an2021neighbor},$_{\!}$~while$_{\!}$~the recent ones explore graph neural~networks$_{\!}$~\cite{deng2020evolving,wang2021structured}$_{\!}$ and$_{\!}$ non-local$_{\!}$ transformer$_{\!}$ models$_{\!}$~\cite{hong2020recurrent,pashevich2021episodic,chen2021history} for$_{\!}$ long-term$_{\!}$ planning,$_{\!}$ with$_{\!}$ the$_{\!}$ assist$_{\!}$ of$_{\!}$ environment$_{\!}$ map$_{\!}$ building \cite{zhao2022target,chen2022think,liu2023bird}, cross-modal matching~\cite{wang2022counterfactual,zhu2021diagnosing,wang2023lana}, and multi-modal pre-training \cite{majumdar2020improving,hao2020towards,qiao2022hop,qiao2023hop+,an2022bevbert} techniques. Besides the advance in model design, several more challenging VLN datasets, such as R4R \cite{jain2019stay}, RxR~\cite{ku2020room}, and REVERIE~\cite{2020REVERIE}, are developed to address long-term navigation~\cite{jain2019stay,ku2020room} and concise instruction guided navigation~\cite{2020REVERIE}.

$_{\!}$These$_{\!}$ remarkable$_{\!}$ efforts$_{\!}$ follow$_{\!}$ the$_{\!}$ seminal$_{\!}$ work$_{\!}$~of R2R to assume the agent can `perfectly' teleport to and from a fixed small set of locations (pre-stored in a sparse navigation graph) in the Matterport3D~\cite{Matterport3D} environment. Although facilitating the evolution and evaluation of VLN algorithms, such discrete task setup is too simplified to cover the practical challenges that a robot would encounter while naviga- ting the real world, such as environment topology acquirement and localization error. In this work, we focus on per-  forming VLN in continuous environments, setup by~\cite{krantz2020navgraph}.

\sssection{VLN$_{\!}$ in$_{\!}$ Continuous$_{\!}$ Environments$_{\!}$ (VLN-CE).$_{\!}$} Krantz$_{\!}$ \etal \cite{krantz2020navgraph} lift$_{\!}$ the$_{\!}$ discrete$_{\!}$ R2R$_{\!}$ VLN$_{\!}$ task$_{\!}$ setup$_{\!}$ to$_{\!}$ the$_{\!}$ conti-nuous$_{\!}$ setting$_{\!}$ ---$_{\!}$ VLN-CE,$_{\!}$ where$_{\!}$ the$_{\!}$ embodied agent is initiated in freely traversable 3D environments and must execute low-level actions to follow natural language navigation$_{\!}$ instructions. After throwing away the unrealistic assumption of the navigation graph, the VLN task becomes more challenging and closer to the real-world. Later approaches attempt to reproduce the success of the abstract VLN in the continuous counterpart, by building a high-level~action space based on online prediction of waypoints \cite{raychaudhuri2021language,hong2022bridging,krantz2021waypoint,wang2022towards,an2023etpnav}. More recently, \cite{krantz2022sim} explores transferring pre-trained VLN policies to continuous environments, demonstrating advantages over training VLN-CE policies from scratch.

$_{\!}$Our$_{\!}$ agent$_{\!}$ is$_{\!}$ favored$_{\!}$ due$_{\!}$ to$_{\!}$ i)$_{\!}$ its$_{\!}$ model-based$_{\!}$ nature,$_{\!}$ and ii)$_{\!}$~the$_{\!}$ ability$_{\!}$ of$_{\!}$ strategic$_{\!}$ and$_{\!}$ interpretable$_{\!}$ planning.$_{\!}$ First,$_{\!}$~our agent learns and builds an explicit model of the world, freeing navigation policy learning from environment modeling. Second, by visualizing possible futures, our agent is able to develop$_{\!}$ advanced$_{\!}$ plan$_{\!}$ before$_{\!}$ moving$_{\!}$ and$_{\!}$ explain$_{\!}$ its$_{\!}$ beha-
viors.$_{\!}$ In$_{\!}$ contrast,$_{\!}$ current$_{\!}$ agents$_{\!}$ make$_{\!}$ only$_{\!}$ greedy$_{\!}$ decisions
 purely relying on their latent state. Their planning ability is rather weak and the decision mode is non-transparent.

\sssection{World$_{\!}$ Model.$_{\!}$} Equipping$_{\!}$ robot$_{\!}$ machines$_{\!}$ with$_{\!}$ the$_{\!}$ ability$_{\!}$ to build$_{\!}$ world$_{\!}$ models$_{\!}$ is$_{\!}$ viewed$_{\!}$ as$_{\!}$ a$_{\!}$ key$_{\!}$ step$_{\!}$ towards$_{\!}$ the$_{\!}$ next-generation$_{\!}$ of$_{\!}$ AI$_{\!}$~\cite{johnson1983mental,lin2020improving}.$_{\!}$ Capturing high-level aspects of the environment, world models help enable AI agents~for rea- sonable decision-making through simulation (a.k.a., imagination) of possible futures. Towards this direction, a bunch\\ \noindent  of approaches have been recently developed to predict forward dynamics/representations \cite{oh2015action,karl2017deep,chiappa2017recurrent,gregor2018temporal,ke2019learning,li2019walking}, perform sampling-based planning~\cite{sekar2020planning,piche2018probabilistic,yen2020experience}, conduct self-simulation based policy learning~\cite{watter2015embed,wang2020exploring,ha2018world}, and recon- cile the advantages of model-based and model-free RL re- gimes \cite{racaniere2017imagination,nagabandi2018neural,srinivas2018universal,lee2020stochastic}. Some others leverage world models$_{\!}$ to anticipate$_{\!}$ (short-term)$_{\!}$ targets$_{\!}$ for$_{\!}$ goal-conditioned$_{\!}$ po- licy$_{\!}$ learning$_{\!}$~\cite{nair2020goal,nair2018visual},$_{\!}$ boost$_{\!}$ the$_{\!}$ learning$_{\!}$ of$_{\!}$ world$_{\!}$ models$_{\!}$~by$_{\!}$ 
introducing$_{\!}$ belief$_{\!}$ of$_{\!}$ state$_{\!}$ uncertainty$_{\!}$~\cite{chua2018deep,gregor2018temporal,gregor2019shaping},$_{\!}$ informa- tive$_{\!}$ state representation$_{\!}$~\cite{lee2020predictive,hafner2020mastering,stooke2021decoupling},$_{\!}$ and$_{\!}$ model$_{\!}$ regulariza- tion$_{\!}$~\cite{lin2020improving,creswell2021unsupervised,kim2020active}.$_{\!}$ Although$_{\!}$ world$_{\!}$ models$_{\!}$ find$_{\!}$ successful$_{\!}$ ap- plications$_{\!}$ in$_{\!}$ Atari~\cite{kaiser2019model,hafner2019learning,hafner2019dream,hafner2020mastering} and robot motion planning \cite{finn2017deep,rybkin2018learning}, many of them are restricted to relatively simple tasks or low-dimensional environments, pointed out by~\cite{hafner2019dream,hafner2020mastering,nair2020goal,koh2021pathdreamer}. To date, world models have been rarely investigated in visually-rich,$_{\!}$ embodied$_{\!}$ tasks.$_{\!}$ The$_{\!}$ few$_{\!}$ excep-

\noindent tions$_{\!}$~\cite{li2019walking,koh2021pathdreamer} simply$_{\!}$ use$_{\!}$ a$_{\!}$ viewpoint$_{\!}$ synthesis$_{\!}$ network$_{\!}$ as$_{\!}$ the$_{\!}$ world$_{\!}$ model. They$_{\!}$ either$_{\!}$ pre-access$_{\!}$ the$_{\!}$ entire$_{\!}$ environ- ment$_{\!}$ typology$_{\!}$~\cite{koh2021pathdreamer}, or only treat the synthesized observations as sub-goals without sampling-based planning~\cite{li2019walking}.$_{\!}$ In~\cite{adamkiewicz2022vision}, NeRF-based world model is adopted for drone navigation planning, which requires pre-exploration of the environment. For$_{\!}$ ours, in$_{\!}$ addition~to$_{\!}$ the$_{\!}$ challenging$_{\!}$ task$_{\!}$ setup, our$_{\!}$ world$_{\!}$ model$_{\!}$ is$_{\!}$ organized$_{\!}$ as$_{\!}$ a$_{\!}$ structured$_{\!}$ summarization$_{\!}$ of$_{\!}$ environment$_{\!}$ layout$_{\!}$ and$_{\!}$ a$_{\!}$ compact representation of environment evolution, instead of \cite{li2019walking,koh2021pathdreamer} encoding all the knowledge about the environment into latent network parameters. More$_{\!}$ importantly, our$_{\!}$ explicit$_{\!}$ and$_{\!}$ discrete$_{\!}$ abstraction$_{\!}$ of the continuous$_{\!}$ environments$_{\!}$ offers$_{\!}$ a$_{\!}$ suitable$_{\!}$ testbed$_{\!}$ for$_{\!}$ mental experiments. Through sampling and assessing numerous possible$_{\!}$ solutions$_{\!}$ entirely$_{\!}$ in$_{\!}$ the$_{\!}$ mind, our$_{\!}$ agent$_{\!}$ can conduct strategic and global navigation planning before moving.

\sssection{Monte Carlo Tree Search (MCTS).} MCTS~\cite{kocsis2006bandit} is a heu- ristic search algorithm for approximating optimal choices~in large$_{\!}$ search$_{\!}$ spaces.$_{\!}$ It$_{\!}$ has$_{\!}$ demonstrated$_{\!}$ great$_{\!}$ success$_{\!}$ in$_{\!}$ creating game-playing$_{\!}$ AI$_{\!}$ agents$_{\!}$ and$_{\!}$ solving$_{\!}$ sequential$_{\!}$ decision$_{\!}$ tasks,$_{\!}$~such as
Atari games~\cite{guo2014deep,schrittwieser2020mastering} and Go~\cite{silver2016mastering,silver2017mastering,schrittwieser2020mastering}. 

In this work, we adopt MCTS to search for possible$_{\!}$~na- vigation$_{\!}$ plans$_{\!}$ based$_{\!}$ on$_{\!}$ the$_{\!}$ world$_{\!}$ model.$_{\!}$ MCTS$_{\!}$~has$_{\!}$~so far been rarely explored in the field of VLN-CE. In$_{\!}$ addition,$_{\!}$~in many$_{\!}$ application$_{\!}$ scenarios$_{\!}$ of$_{\!}$ MCTS,$_{\!}$ all$_{\!}$ the$_{\!}$ aspects$_{\!}$ of$_{\!}$ the world states are fully observable. However, the continuous VLN-CE environments are partially observable, and hence we run MTCS in a discrete, synthesized world space.

\section{Methodology}
\noindent\textbf{Task Setup and Notations.}
In VLN-CE~\cite{krantz2020navgraph}, AI agents~are required to navigate in never-before-seen 3D environments to target positions, according to language instructions. The environments are assumed to be continuous open space. At each step, the agent must choose a low-level action from $\{\textsf{turn-left}~15\degree, \textsf{turn-right}~15\degree, \textsf{move-forward}~0.25\text{m}, \textsf{stop}\}$, given the instruction ${X}$ and $360\degree$ panoramic RGBD obser- vation$_{\!}$ ${Y}$.$_{\!}$ The$_{\!}$ navigation$_{\!}$ is$_{\!}$ successful$_{\!}$ only$_{\!}$ if$_{\!}$ the$_{\!}$ agent$_{\!}$ se- lects \textsf{stop} within 3m of the target location within 500 steps.

\begin{figure*}
	\vspace{-6pt}
	\begin{center}
		\includegraphics[width=0.95\linewidth]{fig/fig2}
	\end{center}
	\vspace{-16pt}
	\captionsetup{font=small}
	\caption{\small (a) Top-down view of current navigation, where \protect\includegraphics[scale=0.07,valign=c]{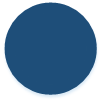} indicates previously
visited waypoints and \protect\includegraphics[scale=0.07,valign=c]{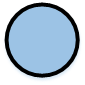} refers to detected but unvisited waypoints. (b) \textsc{Dreamwalker} synthesizes future observations at unvisited waypoints \protect\includegraphics[scale=0.07,valign=c]{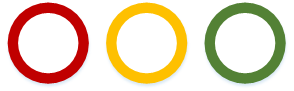} through SS. (c-d) With the synthesized observations, \textsc{Dreamwalker} further extends the synthesized trajectories and looks deeper into the future. (e) \textsc{Dreamwalker} selects
the best mental plan for execution. After reaching the selected waypoint, it starts next-round planning. (f) Top-down view of the waypoint action space. (g) EG $G$ of (a). (h-j) \textsc{Dreamwalker} images its future observation at the unvisited waypoint \protect\includegraphics[scale=0.07,valign=c]{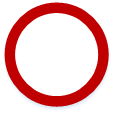}, based on its current observation. The synthesized and real observations at \protect\includegraphics[scale=0.07,valign=c]{fig/o} are given in  (i) and (j), respectively. For clarity, only RGB observation is provided. Notably, the imagined scenarios explain agent's inner decision mode in a way that human can understand, leading to improved interpretability.}
	\label{fig:2}
	\vspace{-10pt}
\end{figure*}

\noindent\textbf{Waypoint$_{\!}$ Action$_{\!}$ Space.$_{\!}$} Recent$_{\!}$ VLN-CE$_{\!}$ solutions$_{\!}$~\cite{raychaudhuri2021language,hong2022bridging,krantz2021waypoint,krantz2022sim}$_{\!}$ adopt$_{\!}$ a$_{\!}$ high-level$_{\!}$ waypoint$_{\!}$ action$_{\!}$ space.$_{\!}$ During$_{\!}$~na-vigation,$_{\!}$ the$_{\!}$ agent$_{\!}$ uses$_{\!}$ a$_{\!}$ \textsl{Waypoint\_Predictor}$_{\!}$ to get a heat- map of 120 angles-by-12 distances, which highlights navigable waypoints in its surrounding. Each angle is of 3 degrees, and the distances range from 0.25 meters to 3.00 meters with 0.25 meters separation, respectively corresponding to the turning angle and forward step size of the low-level action$_{\!}$ space$_{\!}$ (Fig.$_{\!}$~\ref{fig:2}(f)).$_{\!}$ In$_{\!}$ this$_{\!}$ way,$_{\!}$ the$_{\!}$ problem$_{\!}$ of$_{\!}$ inferring low-level controls is translated to the task of waypoint prediction and selection.  Please refer to~\cite{hong2022bridging} for more details.

\noindent\textbf{Algorithmic Overview.} Our \textsc{Dreamwalker} agent solves VLN-CE through \textbf{i)} building a world model (\S\ref{sec:world_model}) that explicitly abstracts intrinsic  properties of continuous environments into a structured, discrete representation; \textbf{ii)} condu- cting strategic planning in the discrete mental world before taking actual actions in the continuous environment (\S\ref{sec:mental_planning}).

\subsection{World Model: Structured, Discrete, and Compact Abstraction of Continuous Environments}
\label{sec:world_model}
To efficiently plan and act in the highly complex world, humans develop a mental model to represent our knowledge about the surrounding, based on our past daily experiences and current information perceived by limited sense \cite{ha2018world}.  In view$_{\!}$ of$_{\!}$ this,$_{\!}$ our$_{\!}$ world$_{\!}$ model$_{\!}$ is$_{\!}$ built$_{\!}$ as$_{\!}$ combination$_{\!}$ of$_{\!}$~two parts:$_{\!}$ i)$_{\!}$ An$_{\!}$ \textit{environment$_{\!}$ graph}$_{\!}$ (EG) -- a structured and disc- rete$_{\!}$ representation$_{\!}$ of$_{\!}$ agent's$_{\!}$ temporary$_{\!}$ knowledge$_{\!}$ about$_{\!}$ the visual landmarks and typologies of the current partially ob- served$_{\!}$ environment;$_{\!}$ and$_{\!}$ ii)$_{\!}$ A$_{\!}$ \textit{scene$_{\!}$ synthesizer}$_{\!}$ (SS)$_{\!}$ that$_{\!}$~en- codes$_{\!}$ agent's$_{\!}$ stable$_{\!}$ knowledge$_{\!}$ about$_{\!}$ some$_{\!}$ general$_{\!}$ rules$_{\!}$~of environments$_{\!}$ such$_{\!}$ as$_{\!}$ transition$_{\!}$ dynamics$_{\!}$ and$_{\!}$ room$_{\!}$ layout, which are learned from training experiences and utilized to predict the unobserved portions of the current environment.

\noindent\textbf{Environment Graph (EG).} EG is organized as an \textit{episodic graph}  $\mathcal{G}\!=\!(\mathcal{V},\mathcal{E})$, where nodes  $v\!\in\!\mathcal{V}$ denote previously~vi- sited$_{\!}$ waypoints$_{\!}$ and$_{\!}$ edges$_{\!}$ $e_{u,v\!}\!\in\!\mathcal{E}_{\!}$ store$_{\!}$ geometric$_{\!}$ relations
between$_{\!}$ waypoints$_{\!}$ (Fig.$_{\!}$~\ref{fig:2}(a)(g)).$_{\!}$ At$_{\!}$ the$_{\!}$ beginning$_{\!}$ of$_{\!}$ current navigation episode, $\mathcal{G}$ only contains one single node --- the starting location. Then, the agent predicts a set of accessible  waypoints,$_{\!}$ and$_{\!}$ selects$_{\!}$ one$_{\!}$ of$_{\!}$ them$_{\!}$ to$_{\!}$ navigate.$_{\!}$ After$_{\!}$~rea- ching the selected waypoint, $\mathcal{G}$ will be updated by involving this waypoint. Hence EG evolves with the navigation~pro-
ceeds. Concretely, for each node $v\!\in\!\mathcal{V}$, its embedding is the feature of the corresponding waypoint observation:
\vspace{-1pt}
\begin{equation}\small
\begin{aligned}
\bm{v} \!=\!\bm{Y}_{p_v},
    \end{aligned}
        \vspace{-3pt}
\end{equation}
where $p_{v\!}$ refers to the location of waypoint $v$, and the coordinate of the start point is set as $(0,0)$; $\bm{Y}_{p_v\!}$ indicates the embedding of the panoramic observation ${Y}_{p_v\!}$ at location $p_v$.

For each edge $e_{u,v}\!\in\!\mathcal{E}$, its embedding $\bm{e}_{u,v}$ encodes topological relations between waypoints $u$ and $v$:
\vspace{-2pt}
\begin{equation}\small
\begin{aligned}
\bm{e}_{u,v}\!=\!(\cos\theta_{u,v},\sin\theta_{u,v}, \bigtriangleup p_{u,v}),
    \end{aligned}
    \vspace{-2pt}
\end{equation}
where $\theta_{u,v}$ and $\bigtriangleup p_{u,v}=p_{u}-p_{v}$ refer to the angle and dis-

\noindent tance between $u$ and $v$, respectively. Note that the connecti- vity among waypoints is also captured by $\mathcal{G}$. An edge, $\bm{e}_{u,v}$, exists$_{\!}$ only$_{\!}$ if$_{\!}$ $u_{\!}$ and$_{\!}$ $v_{\!}$ are$_{\!}$ connected,$_{\!}$ that$_{\!}$ is,$_{\!}$ they$_{\!}$ are$_{\!}$ detected as valid in the waypoint prediction heatmap of each other.

$_{\!}$As$_{\!}$ such,$_{\!}$ EG$_{\!}$ is$_{\!}$ built$_{\!}$ upon$_{\!}$ the$_{\!}$ information$_{\!}$ gathered$_{\!}$ during current navigation episode. Hence it represents the agent's temporary knowledge about its observed surroundings, and organizes$_{\!}$ them$_{\!}$ in$_{\!}$ a$_{\!}$ concise,$_{\!}$ structured,$_{\!}$ and$_{\!}$ discrete$_{\!}$ manner.

\noindent\textbf{Scene$_{\!}$ Synthesizer$_{\!}$ (SS).$_{\!}$} SS$_{\!}$ is$_{\!}$ a$_{\!}$ generative$_{\!}$ network$_{\!}$ that$_{\!}$~pre-
dicts future scenes based on agent's past observations~only,
 without having to navigate them (Fig.$_{\!}$~\ref{fig:2}(h-j)). Given pano-
 ramic RGBD observation $Y_p$ perceived at position $p$, SS is to render a plausible, full-resolution panoramic RGBD observation $\hat{Y}_{p'}$ at an unvisited position $p'$. The position $p'$ of interest is a waypoint of $p$, thus $p$ and $p'$ are spatially close. Like previous world structure-aware video synthesis methods~\cite{mallya2020world,koh2021pathdreamer}, we first project $Y_p$ to
 a 3D point cloud using the depth information and re-project this point cloud into the observation space at position $p'$, so as to obtain a sparse, geometry-aligned RGBD panoramic image ${Y}_{p\rightarrow p'}$. The SS network takes as inputs both the observation $Y_p$ perceived at $p$ and the reconstructed observation ${Y}_{p\rightarrow p'}$, and synthesizes the observation $\hat{Y}_{p'}$ for the unvisited waypoint position $p'$:
 \vspace{-2pt}
\begin{equation}\small
\begin{aligned}
 \hat{Y}_{p'} = \textsl{Scene\_Synthesizer}(Y_p, {Y}_{p\rightarrow p'}).
    \end{aligned}
    \vspace{-2pt}
\end{equation}
SS$_{\!}$ learns$_{\!}$ from$_{\!}$ experiences$_{\!}$ of$_{\!}$ `viewing'$_{\!}$ numerous$_{\!\!}$ VLN-CE training$_{\!}$ environments.$_{\!}$ Its$_{\!}$ parameters$_{\!}$ encode$_{\!}$ agent's$_{\!}$~statis- tical knowledge about some fundamental constraints in the world, such as the transition dynamics and general rules behind room arrangement (\eg, `\textit{dishwasher is typically located in the kitchen}').

\noindent\textbf{Working$_{\!}$ Mode$_{\!}$ of$_{\!}$ World$_{\!}$ Model.$_{\!}$} Integrating$_{\!}$ EG$_{\!}$ and$_{\!}$ SS together$_{\!}$ leads$_{\!}$ to$_{\!}$ a$_{\!}$ powerful$_{\!}$ world$_{\!}$ model.$_{\!}$ EG$_{\!}$ stores$_{\!}$ agent's understanding of the  observed  portions  of the environment.  Based$_{\!}$ on$_{\!}$
 working$_{\!}$ memory,$_{\!}$ SS$_{\!}$ makes$_{\!}$ use$_{\!}$ of$_{\!}$ statistical$_{\!}$ know- ledge to forecast the unexplored portion of the environment  (Fig.$_{\!}$~\ref{fig:2}(a-e)). For instance, starting from a waypoint $v$ in~$\mathcal{G}$,

\noindent the world model uses SS to imagine the future (\ie, $\hat{Y}_{p_{v'}}$) if the agent navigates to a previously unvisited waypoint $v'$ of $v$. With the synthesized observation $\hat{V}_{p_{v'}}$ at waypoint $v'$, the world model can make further future prediction. Notably, our world model, or more precisely, its parametric part -- SS, is trained independently, which lifts the navigation policy from learning inherent structures  of environments.

\subsection{Mental Planning: Forecasting the Future in the World Model}
\label{sec:mental_planning}
Powered by the world model,  \textsc{Dreamwalker} gains the ability of anticipating the consequences of its actions, for an extended$_{\!}$ period$_{\!}$ into$_{\!}$ the$_{\!}$ future.$_{\!}$ This$_{\!}$ is$_{\!}$ because$_{\!}$ the$_{\!}$ world$_{\!}$~mo- del can simulate how the environment changes in response to$_{\!}$ agent's$_{\!}$ actions.$_{\!}$ \textsc{Dreamwalker}$_{\!}$ can$_{\!}$ thus$_{\!}$ make$_{\!}$ advanced planning, through perform mental experiments in the simu- lated world. Basically, at every navigation-decision making step, \textsc{Dreamwalker} first uses the world model to synthesize many future navigation trajectories, and then selects the best one for execution. Here we adopt Monte Carlo Tree Search (MCTS)~\cite{kocsis2006bandit,guo2014deep}, a powerful approach for decision problems, to achieve world model based online planning.

\sssection{MCTS based Mental Planning.} As a~best-first tree search algorithm, MCTS represents the space of candidates into a tree and finds an optimal solution in an iterative manner. In our case, each tree node $s$ represents a possible world state: the$_{\!}$ root$_{\!}$ node$_{\!}$ $s_0$ is$_{\!}$ the$_{\!}$ world$_{\!}$ state$_{\!}$ $\mathcal{G}$ at$_{\!}$ current$_{\!}$ navigation step, while other nodes indicate future world states, synthesized$_{\!}$ by$_{\!}$ the$_{\!}$ world$_{\!}$ model.$_{\!}$ The$_{\!}$ edge$_{\!}$ from$_{\!}$ node$_{\!}$ $s$$_{\!}$~to$_{\!}$~$s'{\!}$
 denotes$_{\!}$ the$_{\!}$ action$_{\!}$ $a$ taken$_{\!}$ in$_{\!}$ $s$ to$_{\!}$ reach$_{\!}$ $s'$,$_{\!}$~and is~identified by pair $(s, a)$.

\begin{figure}
	\vspace{-6pt}
	\begin{center}
		\includegraphics[width=1.01\linewidth]{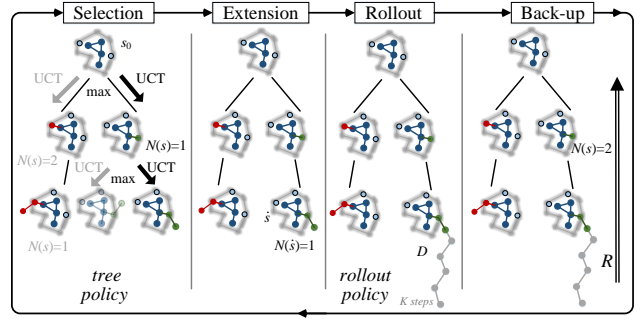}
	\end{center}
	\vspace{-16pt}
	\captionsetup{font=small}
	\caption{\small MCTS based Mental planning. Each node in the research tree refers to a possible world state, corresponding to a future plan. }
	\label{fig:3}
	\vspace{-6pt}
\end{figure}

$_{\!}$The$_{\!}$ core$_{\!}$ idea$_{\!}$ of$_{\!}$ MCTS$_{\!}$ based$_{\!}$ planning$_{\!}$ is$_{\!}$ to$_{\!}$ gradually$_{\!}$ ex- pand the search tree and evaluate rewards, \ie, create many plans and assess the outcomes in mind. This is achieved by an iterative tree search process. Each search iteration starts from the root state $s_0$, and sequentially samples states and actions, based on the simulation of four phases (Fig.~\ref{fig:3}):
\begin{itemize}[leftmargin=*]
    \setlength{\itemsep}{0pt}
	\setlength{\parsep}{0pt}
	\setlength{\parskip}{0pt}
	\setlength{\leftmargin}{-10pt}
   \item[i)] \textit{Selection}: Actions/edges are selected according to a \textit{tree policy}. A commonly used policy is to select greedily with respect to Upper Confidence bounds for Trees (UCT)~\cite{auer2002finite}:	\vspace{-4pt}
      \begin{equation}\small
   \begin{aligned}
     \text{UCT}(s,a) = Q(s,a) + C\sqrt{\frac{\log N(s)}{N(s,a)}},
   \label{equ:uct}
   \end{aligned}
      \end{equation}
   where $Q(s,a)$ is the average accumulated reward of taking action $a$; $N(s, a)$ and $N(s)$ return the visiting times of edge $a$ and state $s$, respectively; $C$ is a scalar explora- tion$_{\!}$ constant.$_{\!}$
   Starting from the root node, the action is selected~as:	\vspace{-3pt}
   \begin{equation}\small
    a^*=\argmax_{a\in \mathcal{A}(s)}\text{UCT}(s,a),
   \end{equation}
   where $\mathcal{A}(s)$ is the global action space at state $s$, which enumerates$_{\!}$ all$_{\!}$ the$_{\!}$ possible$_{\!}$ waypoints.$_{\!}$
    The$_{\!}$ selection$_{\!}$ is$_{\!}$ performed recursively until an unexpanded edge is selected.
   \item[ii)] \textit{Expansion}: With the finally selected unexpanded edge, a new leaf node $\dot{s}$ is further appended, \ie, $N(\dot{s})=1$.
   \item[iii)] \textit{Rollout}: A quick rollout is performed to predict further multiple steps into the future, according to a certain \textit{rollout policy}. Based on a reward function $R$, the value of the new leaf node $\dot{s}$, \ie, $V(\dot{s})$, is obtained.
   \item[iv)] \textit{Back-up}: The statistics of nodes and edges are updated bottom-up through the tree from $\dot{s}$ until reaching the root node.
   \begin{equation}\small
    \begin{aligned}   \label{equ:6}
      N(s,a) &= N(s'),\\
      Q(s,a) &= R(s,a) + \gamma V(s'), \\
      N(s) &= 1+\textstyle\sum\nolimits_{a\in \mathcal{A}(s)}N(s,a), \\
       V(s) &= \textstyle\frac{1}{N(s)}\sum_{a\in \mathcal{A}(s)}\big(N(s,a)Q(s,a)\big).
    \end{aligned}
   \end{equation}
   Here $R(s,a)$ gives the reward for taking action $a$ at state
   $s$; $s'\!=\!T(s,a)$, where $T(s,a)$ is the deterministic transi- tion function.
    The design of our reward function $R$ and the {rollout policy} will be detailed later.
 \end{itemize}
	\vspace{-3pt}
As such, in each iteration, \textsc{Dreamwalker} creates and executes a plan in the world model by expanding the search tree, and anticipates the consequence by estimating the rewards. After several times of iterative simulation, the agent conducts a lot of mental experiments, and the best action is selected among the edges starting from the root node:
	\vspace{-3pt}
 \begin{equation}\small
 \begin{aligned}
  a^*=\mathop{\arg \max}_{a\in\mathcal{A}(s_0)} Q(s_0, a).
 \end{aligned} \vspace{-3pt}
 \end{equation}	

Here$_{\!}$ $a^{*\!}$ corresponds$_{\!}$ to$_{\!}$ a$_{\!}$ certain$_{\!}$ waypoint$_{\!}$ observed$_{\!}$ at$_{\!}$ current world state $s_0$. According to EG $G$, the agent can reach this waypoint by taking a sequence of low-level actions. If $a^*$ is never visited before, EG is updated after its first visit, followed by new-round mental planning and next-step action. When $a^*$ is a visited waypoint, the agent will move to this waypoint and choose \textsf{stop} to terminate navigation.~In~this way,$_{\!}$ our$_{\!}$ agent$_{\!}$ predicts$_{\!}$ future$_{\!}$ in$_{\!}$ the$_{\!}$ discrete$_{\!}$ world$_{\!}$ model, and navigates in the continuous environment.

\sssection{Reward.} The immediate reward $R(s, a)$ is defined according to the change of distance to goal after taking action $a$. Let $\mathcal{G}_{s\!}\!=\!(\mathcal{V}_s,\mathcal{E}_s)$ denote EG corresponding to state $s$, where $\mathcal{V}_s\!-\!\mathcal{V}$ refers to those waypoints visited in dream world state $s$, the distance to the target location in state $s$ is defined as:
\vspace{-3pt}
   \begin{equation}\small
    \begin{aligned}
   D(s)=\min_{v\in\mathcal{V}_s}{F}_d(v, \mathcal{G}_s,{X}),
   \label{equ:distance}
    \end{aligned}
       \vspace{-2pt}
   \end{equation}
where ${F}_d$ is a learnable distance function that predicts the distance from waypoint $v$ to the target location, conditioned on$_{\!}$ EG$_{\!}$ $\mathcal{G}_{s\!}$ and$_{\!}$ instruction$_{\!}$ $X$.$_{\!}$ More$_{\!}$ specifically,$_{\!}$  $D(s)_{\!}$ is$_{\!}$ built upon a graph attention (GAT) network~\cite{velivckovic2018graph}:
       \vspace{-3pt}
  \begin{equation}\small
    \begin{aligned}\label{equ:9}
      {F}_d(v, \mathcal{G}_s,{X}) &= \text{MLP}(\tilde{\bm{v}}),\\
      \tilde{\bm{V}} = \{\tilde{\bm{v}}\}_{v} &= \text{GAT}(\{[\bm{v}, \bm{X}]\}_{v}, \{[\bm{e}_{u,v},\bm{u}, \bm{X}]\}_{u,v}),
    \end{aligned}
           \vspace{-2pt}
   \end{equation}
 where $\bm{v}$ ($\bm{u}$) is the initial embedding of node $v$ ($u$), $\bm{e}_{u,v}$ is the edge embedding of $e_{u,v}\!\in\!\mathcal{E}$. 
 $\bm{X}$ denotes the textual embedding of the instruction $X$, and $[\cdot,\cdot]$ refers to concatenation. After fusing node and edge embeddings with the textual context, GAT outputs the collection of improved
 node embeddings $\tilde{\bm{V}}$. After that, a small multilayer perceptron (MLP) is applied per node for distance regression.

Given the deterministic transition $s'\!=\!T(s,a)$, we have $a\!=\!\mathcal{V}_{s'\!}\!-\!\mathcal{V}_s$. If action $a$ is a previously visited waypoint, \ie, $a\!\in\!\mathcal{V}$, the immediate reward $R(s,a)$ is given as:
 \begin{equation}\small
  R(s,a)\!=\!\left\{
    \begin{aligned}
      +5, \quad {F}_d(v, \mathcal{G}_s,{X})\leq 3,\\
      -5, \quad {F}_d(v, \mathcal{G}_s,{X})> 3.
    \end{aligned}
  \right.
 \end{equation}
Here$_{\!}$ $a\!\in\!\mathcal{V}_{\!}$ means$_{\!}$ the$_{\!}$ agent$_{\!}$ chooses$_{\!}$ \textsf{stop}$_{\!}$ at$_{\!}$ waypoint$_{\!}$ $a$.$_{\!\!}$ We estimate if its distance to the target is within the success~cri- teria,$_{\!}$ \ie,$_{\!}$ 3m,$_{\!}$ and$_{\!}$ assign$_{\!}$ a$_{\!}$ constant$_{\!}$ positive/negative$_{\!}$ reward (+5/-5)$_{\!}$ accordingly.$_{\!}$ If$_{\!}$ waypoint$_{\!}$ $a$$_{\!}$ is$_{\!}$ never$_{\!}$ visited$_{\!}$ before, we~define:
   \begin{equation}\small
    \begin{aligned}\label{equ:11}
  R(s,a)\!=D(s')-D(s).
    \end{aligned}
   \end{equation}
That is to say, the reward is defined as the distance that $a$ can bring the agent closer to the target than before.

\sssection{Rollout Policy.} In the rollout phase, a rollout policy~is~ado- pted$_{\!}$ to$_{\!}$ guide$_{\!}$ the$_{\!}$ fast$_{\!}$ playout$_{\!}$ starting$_{\!}$ from$_{\!}$ the$_{\!}$ new$_{\!}$ expanded leaf state $\dot{s}$. This can be intuitively viewed as further imagining the future of state $\dot{s}$ with several steps through fast~sa- mpling.$_{\!}$ For$_{\!}$ the$_{\!}$ sake$_{\!}$ of$_{\!}$ Monte$_{\!}$ Carlo$_{\!}$ property$_{\!}$ and$_{\!}$ simplicity, we treat the distance function $D(\cdot)$ (\textit{cf}.~Eq.~\ref{equ:distance}) as the rollout policy. 
For example, at the first rollout step, the action distribution $\bm{p}$ over possible waypoints $\mathcal{A}(\dot{s})$ at state $\dot{s}$ is given as:
\begin{equation}\small
\begin{aligned}
  \bm{p}[a]=\text{softmax}_{a\in\mathcal{A}(\dot{s})}{F}_d(a, \mathcal{G}_{\dot{s}},{X}).
\end{aligned}
\vspace{3pt}
\end{equation}
The rollout stops when a previously visited waypoint is selected or a maximum rollout depth $K$ is reached. The rewards $\{R_1,R_2,\cdots,R_K\}$ along the rollout record are collected to compute the accumulated discounted reward as the value of $\dot{s}$:
\vspace{-1pt}
\begin{equation}\small
\begin{aligned}\label{equ:13}
  V(\dot{s})=\textstyle\sum\nolimits_{k=1}^K\gamma^{k-1}R_k.
\end{aligned}
\end{equation}

\section{Experiments}
After stating our experimental setup (\S\ref{sec:es}) and implementation details (\S\ref{sec:id}), we provide performance comparison results with VLN-CE state-of-the-arts (\S\ref{sec:exm}). Then we identify there is still room for improvement in the aspects of world model (\S\ref{sec:gwm}) and distance estimation (\S\ref{sec:idf}), revealing possible future directions. We further verify the contribution of our world model based mental planning (\S\ref{sec:ap}). Finally, we evaluate the impacts of core hyper-parameters (\S\ref{sec:dex}) and offer visual analyses (\S\ref{sec:qr}).

\subsection{Experimental Setup}\label{sec:es}
\sssection{Dataset.} We conduct experiments on VLN-CE dataset~\cite{krantz2020navgraph}. The$_{\!}$ dataset$_{\!}$ has$_{\!}$ $16,844_{\!}$ trajectory-instruction$_{\!}$ pairs$_{\!}$ across $90_{\!}$ Matterport3D$_{\!}$~\cite{Matterport3D} scenes, and is divided into four sets, \ie, \texttt{train} ($10,819$ pairs,$_{\!}$ $61_{\!}$ scenes),$_{\!}$ \texttt{val}$_{\!}$ \texttt{seen}$_{\!}$ ($778_{\!}$ pairs,$_{\!}$ $53_{\!}$ scenes), \texttt{val} \texttt{unseen} ($1,839$ pairs, $11$ scenes), and \texttt{test} ($3,408$ pairs,$_{\!}$ $18_{\!}$ scenes).$_{\!}$ As$_{\!}$ the$_{\!}$ scenes$_{\!}$ in$_{\!}$ \texttt{val}$_{\!}$ \texttt{unseen}$_{\!}$ and \texttt{test} are not exposed in \texttt{train}, the performances on
\texttt{val} \texttt{unseen} and \texttt{test} are more important than \texttt{val} \texttt{seen}.

\sssection{Evaluation Metric.} Following \cite{krantz2020navgraph,krantz2021waypoint,krantz2022sim}, we use five me- trics$_{\!}$ for$_{\!}$ evaluation,$_{\!}$ \ie,$_{\!}$ Navigation$_{\!}$ Error$_{\!}$ (NE),$_{\!}$ Trajectory Length (TL), Success Rate (\textbf{SR}), Oracle success Rate (OR), and Success rate weighted by Path Length (SPL), where {SR} is the priority. Please see \cite{anderson2018evaluation,anderson2018vision} for full details on metrics.

\subsection{Implementation Details}\label{sec:id}

\sssection{Network$_{\!}$ Architecture.$_{\!}$} As$_{\!}$ in$_{\!}$~\cite{krantz2021waypoint,krantz2022sim,krantz2020navgraph,hong2022bridging},$_{\!}$ the$_{\!}$ RGB$_{\!}$ and$_{\!}$ depth$_{\!}$ channels$_{\!}$ of$_{\!}$ the$_{\!}$ panoramic$_{\!}$ observation$_{\!}$ $Y_{\!}$ are$_{\!}$ respec-  tively$_{\!}$ encoded$_{\!}$ by$_{\!}$ ImageNet$_{\!}$~\cite{ILSVRC15}$_{\!}$ pre-trained$_{\!}$ ResNet-18$_{\!}$~\cite{he2016deep} and$_{\!}$ PointGoal$_{\!}$~\cite{wijmans2019dd}$_{\!}$ pre-trained$_{\!}$ ResNet-50.$_{\!\!}$ The$_{\!}$ language$_{\!}$~ins-
truction $X$ is encoded through a bi-LSTM with GLoVE$_{\!}$~\cite{jeffreypennington2014glove} word embeddings. \textsl{Scene\_Synthesizer} is built as a two-stage generator,$_{\!}$ following$_{\!}$~\cite{koh2021pathdreamer}.$_{\!\!}$ \textsl{Waypoint\_Predictor}$_{\!}$ is$_{\!}$ the$_{\!}$ one$_{\!}$ used$_{\!}$ in$_{\!}$~\cite{hong2022bridging}. For the distance function $F_d$, GAT is implemented as standard~\cite{velivckovic2018graph}, and MLP has $1,024$ hidden neurons.

\sssection{Network$_{\!}$ Training.$_{\!}$} The$_{\!}$ training$_{\!}$ of$_{\!\!}$ \textsl{Waypoint\_Predictor}$_{\!}$~and \textsl{Scene\_Synthesizer} are scheduled as in \cite{hong2022bridging} and \cite{koh2021pathdreamer} respec- tively.$_{\!}$ For our$_{\!}$ distance$_{\!}$ function$_{\!}$ $F_d$,$_{\!}$ we progressively construct EGs along the ground-truth trajectory of each training episode. $F_d$ is trained by minimizing the L2 loss between the predicted distance and ground-truth distance for each waypoint for each EG. More specifically, for robust distance prediction, for a ground-truth trajectory with $L$ waypoints, we construct $L$ EGs where the $l$-th EG is constructed by adding the $l$-th waypoint of the trajectory as well as up to 5 random sampled, accessible waypoints into the $(l\!-\!1)$-th EG.  In addition, we randomly replace ground-truth waypoints with the ones created by \textsl{Scene\_Synthesizer}, making $F_d$ better adapted to the working mode of mental planning.

\sssection{Reproducibility.} Our hyper-parameters are set as follows. The$_{\!}$ discount$_{\!}$ factor$_{\!}$ $\gamma_{\!}$ is$_{\!}$ $0.98_{\!}$ (Eq.$_{\!}$~\ref{equ:6}$_{\!}$ and$_{\!}$~\ref{equ:13}).$_{\!}$ The$_{\!}$ horizon$_{\!}$~of mental planning, \ie, the maximum step of imagining the future$_{\!}$ trajectory,$_{\!}$ is$_{\!}$ $4$.$_{\!}$ The$_{\!}$ number$_{\!}$ of$_{\!}$ search$_{\!}$ iterations$_{\!}$~for each time of mental planning is $50$. Our model is implemented in PyTorch with Habitat~\cite{2019Habitat} simulator, and trained on one NVIDIA RTX 3090 GPU.
Our code is released at \url{https://github.com/hanqingwangai/Dreamwalker}.

\subsection{Comparison with VLN-CE State-of-the-Arts}\label{sec:exm}

\begin{figure*}
\begin{minipage}{\textwidth}
\begin{minipage}[t]{0.72\textwidth}
    \begin{threeparttable}
        \resizebox{0.98\textwidth}{!}{
		\setlength\tabcolsep{2pt}
            \renewcommand\arraystretch{1.03}
    \begin{tabular}{rl||cc>{\columncolor[gray]{0.90}}ccc|cc>{\columncolor[gray]{0.90}}ccc|cc>{\columncolor[gray]{0.90}}ccc}
  \hline \thickhline
  \rowcolor{mygray}
   ~ & & \multicolumn{5}{c|}{\texttt{val} \texttt{seen}}& \multicolumn{5}{c|}{\texttt{val} \texttt{unseen}}& \multicolumn{5}{c}{\texttt{test}} \\
  \cline{3-17}\cline{3-17}\cline{3-17}\cline{3-17}
  \rowcolor{mygray}
  \multicolumn{2}{c||}{\multirow{-2}{*}{Model}} &NE$\downarrow$ &TL &SR$\uparrow$ & OR$\uparrow$ &SPL$\uparrow$ &NE$\downarrow$ &TL &SR$\uparrow$ & OR$\uparrow$ & SPL$\uparrow$&NE$\downarrow$ &TL &SR $\uparrow$ & OR$\uparrow$ & SPL$\uparrow$\\
  \hline
  \hline
   CMA~\cite{krantz2020navgraph}\!\!&\pub{ECCV20}& 7.21 & 9.06 & 34 & 44 & 32 & 7.60 & 8.27 & 29 & 36 & 27 & 7.91 & 8.85 & 28 & 36 & 25 \\
   Waypoint~\cite{krantz2021waypoint}\!\!&\pub{ICCV21} & 5.48 & 8.54 & 46 & 53 &43 & 6.31 & 7.62 & 36 & 40 & 34 & 6.65 & 8.02 & 32 & 37 & 30\\
   LAW~\cite{raychaudhuri2021language}\!\!&\pub{EMNLP21} &6.35 &9.34 &40 &49 &37 &6.83 &8.89  &35 &44 &31 &7.69  &9.67 &28 &38 &25 \\
   BridgingGap~\cite{hong2022bridging}\!\!&\pub{CVPR22} & 5.02 & 12.5 & 50 & 59 & 44 & 5.74 & 12.2 & 44 & 53 & 39 & 5.89 & 13.3 & 42 & 51 & 36\\
   Sim2Sim~\cite{krantz2022sim}\!\!&\pub{ECCV22}&4.67 &11.2  &52 &61 &44 &6.07 &10.7 &43 &52 &36 &6.17 &11.4 &44 &52 &37\\
  \hline
   \multicolumn{2}{c||}{\textsc{Dreamwalker} (\textbf{Ours})} & \textbf{4.09} & 11.6 & \textbf{59} & \textbf{66} & \textbf{48} & \textbf{5.53} & 11.3 & \textbf{49} & \textbf{59} & \textbf{44} & \textbf{5.48} & 11.8 & \textbf{49} & \textbf{57} & \textbf{44}\\
  \hline
  \end{tabular}
    }
\end{threeparttable}
    \vspace*{-8pt}
    \makeatletter\def\@captype{table}\makeatother\captionsetup{font=small}\caption{\small{$_{\!}$Impacts$_{\!}$ of$_{\!}$ core$_{\!}$ method$_{\!}$ components$_{\!}$ on$_{\!}$ VLN-CE$_{\!}$ dataset$_{\!}$~\cite{krantz2020navgraph}$_{\!}$ (\S\ref{sec:gwm}-\S\ref{sec:ap}). \label{table:mainex}
 }  }
  \end{minipage}
  \begin{minipage}[t]{0.275\textwidth}
   	\vspace*{-2pt}
      \begin{center}
        \includegraphics[width=\linewidth]{fig/curve1}
        \captionsetup{font=small}
     \vspace*{-24pt}
  \caption{\small Curves of success rate and distance estimation error (\S\ref{sec:idf}).}
        \label{fig:curve1}
      \end{center}
   \end{minipage}
  \end{minipage}
  \vspace*{-6pt}
\end{figure*}

We$_{\!}$ compare$_{\!}$ our$_{\!}$ agent,$_{\!}$ \textsc{Dreamwalker},$_{\!}$ with$_{\!}$ five$_{\!}$~pre- viously published VLN-CE models~\cite{krantz2020navgraph,krantz2021waypoint,raychaudhuri2021language,hong2022bridging,krantz2022sim}. As illustrated in Table$_{\!}$~\ref{table:mainex}, our agent outperforms all the competitors across all the splits, in terms of SR. In particular, \textsc{Dreamwalker} surpasses BridgingGap$_{\!}$~\cite{hong2022bridging}, the current top-leading VLN-CE model, by$_{\!}$ \textbf{7\%},$_{\!}$ \textbf{5\%},$_{\!}$ and$_{\!}$ \textbf{7\%},$_{\!}$ on$_{\!}$ \texttt{val}$_{\!}$ \texttt{seen},$_{\!}$ \texttt{val}$_{\!}$ \texttt{unseen},$_{\!}$ and \texttt{test}$_{\!}$ splits,$_{\!}$ respectively.$_{\!}$ Since$_{\!}$ all$_{\!}$ the$_{\!}$ involved$_{\!}$ competitors are model-free approaches, our promising results demonstrate the superiority of our world model based algorithmic design, which relieves the burden on the agent to learn the knowledge about the environments during navigation policy training. In addition, we can find that \textsc{dreamwalker} boosts SR score greatly without intro- ducing$_{\!}$ much$_{\!}$ extra$_{\!}$ travel$_{\!}$ expense.$_{\!}$ This$_{\!}$ evidences$_{\!}$ that,$_{\!}$ compared  to  greedily following a sophisticated navigation policy, planning ahead in the mental world enables advanced decision-making with little expense of physical execution. 

\subsection{How Far We Are From a Perfect World Model?}\label{sec:gwm}
The world model allows our \textsc{Dreamwalker} to employ `mental imagery' to perform mental experiments, so as to plan ahead before taking action. The previous experiments
\begin{figure*}
\begin{minipage}{\textwidth}
\begin{minipage}[t]{0.395\textwidth}
   	\vspace*{-2pt}
\noindent  (\textit{cf}.$_{\!}$~\S\ref{sec:exm}$_{\!}$~and$_{\!}$  Table$_{\!}$~\ref{table:mainex})$_{\!}$  primarily$_{\!}$ demonstrated$_{\!}$ the power$_{\!}$ of$_{\!}$ the$_{\!}$ world$_{\!}$ model$_{\!}$ in$_{\!}$ strategic$_{\!}$ navigation$_{\!}$ planning,$_{\!}$ through$_{\!}$ the$_{\!}$ comparison$_{\!}$ with$_{\!}$ existing model-free$_{\!\!}$ VLN-CE$_{\!}$ agents.$_{\!\!}$ To$_{\!}$ help$_{\!}$ highlight$_{\!}$ how$_{\!}$\\
\noindent far we are from a perfect world model, we derive a$_{\!}$ variant$_{\!}$ --$_{\!}$ ``perfect imagination''$_{\!}$ --$_{\!}$ from$_{\!}$ our$_{\!}$~algo- rithm, by replacing the future scenarios forecas-
   \end{minipage}
   \begin{minipage}[t]{0.005\textwidth}
   ~~~~~~
   \end{minipage}
 \begin{minipage}[t]{0.6\textwidth}
    \begin{threeparttable}
        \resizebox{0.98\textwidth}{!}{
		\setlength\tabcolsep{3pt}
            \renewcommand\arraystretch{1.03}
    \begin{tabular}{c||ccccc|ccccc}
    \hline \thickhline
    \rowcolor{mygray}
      & \multicolumn{5}{c|}{\texttt{val} \texttt{seen}}& \multicolumn{5}{c}{\texttt{val} \texttt{unseen}} \\
    \cline{2-11} \cline{2-11} \cline{2-11} \cline{2-11}
      \rowcolor{mygray}
    \multirow{-2}{*}{{Variant}}  & NE $\downarrow$ &TL &SR $\uparrow$ & OR $\uparrow$ & SPL $\uparrow$ &NE $\downarrow$ &TL &SR $\uparrow$ & OR $\uparrow$ & SPL $\uparrow$\\
    \hline
    \hline
     \textsc{Dreamwalker} (\textbf{Ours}) & 4.09 & 11.6 & 59 & 66 & 48 & 5.53 & 11.3 & 49 & 59 & 44\\
        \hline
     Perfect Imagination & 3.75 & 10.8 & 64 & 69 & 60 & 4.88 & 11.1 & 54 & 63 & 49 \\
     Copy Memory  & 7.10 & 13.5 & 35 & 44 & 31 & 7.76 & 13.8 & 27 & 35 & 24 \\
    \hline
    Greedy Selection& 5.22 & 10.5 & 47 & 56 & 43& 5.93 & 10.9 & 42 & 53 & 36\\
    \hline
    \end{tabular}
    }
\end{threeparttable}
    \vspace*{-8pt}
    \makeatletter\def\@captype{table}\makeatother\captionsetup{font=small}\caption{\small{$_{\!}$Impacts$_{\!}$ of$_{\!}$ core$_{\!}$ method$_{\!}$ components$_{\!}$ on$_{\!}$ VLN-CE$_{\!}$ dataset$_{\!}$~\cite{krantz2020navgraph}$_{\!}$ (\S\ref{sec:gwm}-\S\ref{sec:ap}). \label{table:world}
 }  }
  \end{minipage}
  \end{minipage}
  \vspace*{-20pt}
\end{figure*}

\noindent sted by our world model with the corresponding actual ahead observations from the environment. For completeness, we also provide a ``lazy'' world model -- ``copy memory'' -- which simply memorizes all past observations and generates future predictions by simply copying the nearest memory. From Table~\ref{table:world} we can find that, compared with \textsc{Dreamwalker}, ``perfect imagination'' yields solid performance boost. This is because the agent coupled with a perfect world model can make perfect prediction of the$_{\!}$ future.$_{\!}$ This$_{\!}$ also$_{\!}$ suggests$_{\!}$ the$_{\!}$ upperbound$_{\!}$ of$_{\!}$ our$_{\!}$ performance$_{\!}$ \wrt$_{\!\!\!}$ world$_{\!}$ model,$_{\!}$ and$_{\!}$ is$_{\!}$ consistent$_{\!}$ with$_{\!}$ prior$_{\!}$ studies \cite{wang2020active,koh2021pathdreamer} which find that allowing agent to look ahead at actual observations from environments can facilitate decision-making in the abstract setting. At the other extreme, ``copy memory'', not surprisingly, gives the worst results, since the agent does$_{\!}$ not$_{\!}$ make$_{\!}$ any$_{\!}$ imagination$_{\!}$ of$_{\!}$ the$_{\!}$ future.$_{\!}$ This$_{\!}$ verifies again the benefit of world model in navigation planning.

\subsection{Advanced Planning or Greedy Selection?}\label{sec:ap}
\vspace{-1pt}
It$_{\!}$ is$_{\!}$ also$_{\!}$ interesting$_{\!}$ to$_{\!}$ quantify$_{\!}$ the$_{\!}$ contribution$_{\!}$ of$_{\!}$ our$_{\!}$ world model based mental planning. To this end, in Table~\ref{table:world}, we report a baseline - ``greedy selection'': at each decision-making step, the agent greedily selects the waypoint in the view of field for navigation, according to the estimated distance to the target location. This decision-making strategy is adopted by all current VLN-CE agents, yet lacking explicit planning. Our world model enables sampling-based planning --- first assessing the consequences of possible navigation solutions in the mind then taking actual action. From Table~\ref{table:world} we can find that, mental planning exhibit significantly better performance compared with the greedy action selection strategy. This evidences our primary hypothesis that  imagination of possible futures in the internal world abstract is necessary for strategic navigation planning.

\begin{figure*}
\begin{minipage}{\textwidth}
 \begin{minipage}[t]{0.64\textwidth}
  \begin{minipage}[t]{1\textwidth}
    \begin{threeparttable}
        \resizebox{0.98\textwidth}{!}{
		 \setlength\tabcolsep{3pt}
            \renewcommand\arraystretch{1.03}
    \begin{tabular}{c|c||ccccc|ccccc|c}
    \hline \thickhline
    \rowcolor{mygray}
    ~ &Searching  & \multicolumn{5}{c|}{\texttt{val} \texttt{seen}}& \multicolumn{5}{c|}{\texttt{val} \texttt{unseen}} &Runtime\\
    \cline{3-12} \cline{3-12} \cline{3-12}\cline{3-12}
      \rowcolor{mygray}
      \multirow{-2}{*}{\#}&{Iteration} & NE $\downarrow$ &TL  &SR $\uparrow$ & OR $\uparrow$ & SPL $\uparrow$ &NE $\downarrow$ &TL &SR $\uparrow$ & OR $\uparrow$ & SPL $\uparrow$ &{(\textit{s/step}) $\downarrow$}\\
    \hline
    \hline
    1 & 10 & 4.43 & 12.1 & 55 & 63 & 45 & 5.76 & 12.5 & 44 & 55 & 38 & 0.43\\
    2 & 30 & 4.29 & 10.9 & 57 & 65 & 47 & 5.62 & 11.1 & 46 & 57 & 42 & 1.08\\
    3 & 50 & 4.09 & 11.6 & 59 & 66 & 48 & 5.53 & 11.3 & 49 & 59 & 44 & 1.43\\
    4 & 70 & 4.02 & 12.6 & 59 & 67 & 48 & 5.49 & 12.9 & 50 & 60 & 44 & 1.74\\
    \hline
    \end{tabular}
    }
\end{threeparttable}
    \vspace*{-8pt}
    \makeatletter\def\@captype{table}\makeatother\captionsetup{font=small}\caption{\small{Impact of sampling iteration for navigation plan searching (\S\ref{sec:dex}).\label{table:ab2} 
 }  }
 \vspace*{3pt}
 \end{minipage}
 \hspace*{-5pt}
  \begin{minipage}[t]{1\textwidth}
 \begin{threeparttable}
        \resizebox{\textwidth}{!}{
		 \setlength\tabcolsep{3pt}
            \renewcommand\arraystretch{1.03}
    \begin{tabular}{c|c||ccccc|ccccc|c}
    \hline \thickhline
    \rowcolor{mygray}
    ~ &Planning  & \multicolumn{5}{c|}{\texttt{val} \texttt{seen}}& \multicolumn{5}{c|}{\texttt{val} \texttt{unseen}} &Runtime\\
    \cline{3-12} \cline{3-12} \cline{3-12}\cline{3-12}
      \rowcolor{mygray}
      \multirow{-2}{*}{\#}&{Horizon} & NE $\downarrow$ &TL &SR $\uparrow$ & OR $\uparrow$ & SPL $\uparrow$ &NE $\downarrow$ &TL &SR $\uparrow$ & OR $\uparrow$ & SPL $\uparrow$ &{  (\textit{s/step}) $\downarrow$}\\
    \hline
    \hline
    1 & 0 & 5.22 & 10.5 & 47 & 56 & 43& 5.93 & 10.9 & 42 & 53 & 36 & 0.09\\
    2 & 2 & 4.21 & 11.8 & 56 & 64 & 45 & 5.66 & 12.2 & 47 & 57 & 41 & 1.15 \\
    3 & 4 & 4.09 & 11.6 & 59 & 66 & 48 & 5.53 & 11.3 & 49 & 59 & 44 & 1.43 \\
    4 & 6 & 4.18 & 12.5 & 57 & 65 & 44 & 5.59 & 12.9 & 48 & 58 & 41 & 2.05\\
    \hline
    \end{tabular}
    }
\end{threeparttable}
    \vspace*{-8pt}
    \makeatletter\def\@captype{table}\makeatother\captionsetup{font=small}\caption{\small{Impact of planning horizon (\S\ref{sec:dex}). The maximum searching round is 50.\label{table:ab3} 
 }
 }
  \end{minipage}
  \end{minipage}
    \begin{minipage}[t]{0.005\textwidth}
   ~~~~~~
   \end{minipage}
  \begin{minipage}[t]{0.355\textwidth}
   	\vspace*{-4pt}
      \begin{center}
        \includegraphics[width=\linewidth]{fig/curve2}
        \captionsetup{font=small}
     \vspace*{-22pt}
  \caption{\small FID curve of synthesized panoramic view \wrt prediction step (\S\ref{sec:dex}).}
        \label{fig:curve2}
      \end{center}
   \end{minipage}
  \end{minipage}
  \vspace*{-10pt}
\end{figure*}

\vspace{-2pt}
\subsection{What Is the Impact of Distance Function?}\label{sec:idf}
\vspace{-1pt}
During mental planning, \textsc{Dreamwalker} makes use of the distance function $F_d$ (\textit{cf}.~Eq.~\ref{equ:9}) to calculate the immediate reward $R$ (\textit{cf}.~Eq.~\ref{equ:11}) for a certain action. In other words,  \textsc{Dreamwalker} assesses the outcome of mental plans by means of the distance function. We therefore conduct experiments to study the influence of the distance function. Specifically, we report the performance  by randomly replacing the distance estimates with the ground-truth in different probabilities. The performance are plotted in curves in Fig.~\ref{fig:curve1}. We can observe that, when the distance estimation becomes more accurate, \textsc{Dreamwalker} performs better. With perfect distance estimation -- the agent is allowed to access accurate distance between any waypoint and the target goal, 100\% SR can be reached.  We also find that, even when the replacement probability is as low as 20\%, the agent can achieve a rather high 70 SR. It demonstrates the central role of the distance function in our algorithm and, also, suggests a feasible direction for further improvement.

\subsection{Hyper-Parameter Study}\label{sec:dex}
In this section, we examine our hyper-parameter setup.

\sssection{Search Iterations of MCTS.} We first investigate the influence of searching iterations in MCTS based mental planning. Intuitively, with more searching rounds, the search tree has a higher probability to reach a good terminal state and can better estimate the outcome of possible actions, yet,$_{\!}$ at$_{\!}$ the$_{\!}$ expense$_{\!}$ of$_{\!}$ larger$_{\!}$ simulation$_{\!}$ cost.$_{\!}$ Therefore, in addition to estimating the navigation performance with different searching rounds (\ie, 10, 30, 50, 70), we also report statistics for the runtime. As shown in  Table~\ref{table:ab2}, the runtime  grows linearly as the number of searching rounds increases. However, when the number of searching rounds exceeds 50, the performance$_{\!}$ gain$_{\!}$ becomes$_{\!}$ marginal.$_{\!}$ Thus$_{\!}$ we$_{\!}$ finally$_{\!}$ set$_{\!}$ the number of searching rounds to 50 to save the unproductive computation cost. We also stress that our mental planning is very fast which typically  responds within 1.5s.

\sssection{Horizons of Mental Planning.} We report the performance with different planning horizons --- imaging future trajectories with a maximum length of 0, 2, 4, or 6 forward steps. From Table~\ref{table:ab3} we can find that, both the performance and the runtime generally grows as the horizons increased from 0 to 4 steps. Comparing \#3 and \#4 rows, we can find that the performance gain becomes marginal or even negative.
This happens probably due to that the capacity of the world model is overloaded.  As in \cite{koh2021pathdreamer}, we assess the fidelity of our synthesized panoramic RGBD views by computing Fr\'echet Inception Distance (FID)~\cite{heusel2017gans} to the ground-truth scenes. As shown in Fig.~\ref{fig:curve2}, the error between simulated views and corresponding actual observations from real environments accumulates as the trajectory rolls out. Hence the scenarios forecasted over long horizons might be useless or even detrimental to mental planning.

\begin{figure}
  \begin{center}
    \includegraphics[width=\linewidth]{fig/fig4}
  \end{center}
  \vspace*{-10pt}
  \captionsetup{font=small}
  \caption{\small (a) Trajectories {of} an episode navigated by a greedy policy (red) and \textsc{Dreamwalker} (blue). (b) Current panoramic observation. (c) The search tree rooted by the world state of (a). The nodes and edges are painted according to their $V(s)$ and $Q(s,a)$ respectively. (d) Imagined view at waypoint {\color{sblue}{A}}. (e) Imagined view at waypoint {\color{ppink}B}. See \S\ref{sec:qr} for more details.}
  \label{fig:qr}
    \vspace*{-10pt}
\end{figure}

\subsection{Qualitative Result}\label{sec:qr}
To better understand the superior performance of our method and demonstrate the interpretability of the mental planning mechanism, we analyse a challenging qualitative case in Fig.~\ref{fig:qr}. In this case, we visualize the navigation trajectory performed by a greedy policy (\ie, greedily selecting the best waypoint predicted by the distance function) and our \textsc{Dreamwalker}. As seen, given a complicated instruction ``\textit{Walk out $\cdots$ to outside.}'', the agent with the greedy policy soon gets lost after it enters the dining room, while our \textsc{Dreamwalker} manages to reach the target location. For an intuitive comprehension of the planning procedure, here we visualize a part of the search tree. The nodes and edges in the search tree are colored according to their corresponding  $V(s)$ and $Q(s,a)$ values, \ie, \red{red} color indicates high value and \blue{blue} color indicates low value. We can clearly observe that the search tree is split into two branches when the planning proceeds to ``\textit{walk into the dining area}'', \ie, \textsc{Dreamwalker} starts to imagine the outcomes of two possible actions, and the branch of the correct action finally wins with a large margin as the branch of the wrong action receives rather low rewards (deep \blue{blue}). We visualize the imagined future waypoints of the correct action and the wrong action respectively and find that the correct action leads to a room which looks more like a ``\textit{living room}'' mentioned in the instruction, while the wrong action leads to a corridor. This study demonstrates that our \textsc{Dreamwalker} can provide strong interpretability of decision making by conducting tractable planning and presenting intuitive visualization for imagined observations.

\vspace{-3pt}
\section{Conclusion and Discussion}
\vspace{-1pt}
$_{\!}$In$_{\!}$ this$_{\!}$ article,$_{\!}$ we$_{\!}$ devise$_{\!}$ \textsc{Dreamwalker},$_{\!}$ a$_{\!}$ world$_{\!}$ model based$_{\!}$ VLN-CE$_{\!}$ agent.$_{\!}$ Our$_{\!}$ world$_{\!}$ model$_{\!}$ is$_{\!}$ built$_{\!}$ as$_{\!}$ a$_{\!}$ discrete and structured abstraction of the continuous environment, allowing \textsc{Dreamwalker} to synthesize, assess, and interpret possible plans in the mind before taking actual actions. We empirically confirm the superiority of \textsc{Dreamwalker} over existing model-free agents, in terms of performance and interpretability. Building more expressive and versatile world models, and exploring world model based VLN-CE policy learning will be the focus of our future work.


\setcounter{equation}{0}
\setcounter{section}{0}
\setcounter{figure}{0}

\renewcommand\thesection{\Roman{section}}
\renewcommand\thefigure{\Roman{figure}}
\renewcommand\theequation{\Roman{equation}}
\renewcommand\thealgorithm{\Roman{algorithm}}
\renewcommand\thetable{\Roman{table}}
\renewcommand\thepage{S\arabic{page}}

\newpage
\makesupptitle{\textsc{Dreamwalker}: Mental Planning for Continuous Vision-Language Navigation\\\textit{Supplementary Material}}

This document sheds more details of our approach and additional experimental results, organized as follows:
\begin{itemize}
  \setlength{\itemsep}{0pt}
  \setlength{\parsep}{0pt}
  \setlength{\parskip}{0pt}
  \item \S\ref{sec:implement} Implementation Details.
  \item \S\ref{sec:qualitative} Qualitative Results.
  \item \S\ref{sec:discuss} Discussions.
\end{itemize}

\section{Implementation Details}
\label{sec:implement}

\noindent\textbf{Experimental Configurations.} We utilize the observation space adopted in prior works~\cite{hong2022bridging,krantz2021waypoint,krantz2022sim} where the agent perceives 12 single-view RGBD images at horizontal 30 degrees separation for each move. The size of RGB images and depth map is $224\times224$ and $256\times256$ respectively. The vertical FOV of the camera is 90$^\circ$. Following~\cite{krantz2020navgraph,hong2022bridging}, we apply an ImageNet\cite{ILSVRC15}-pretrained ResNet50~\cite{he2016deep} for RGB feature extraction, and a modified ResNet50 pretrained in point-goal navigation~\cite{2019Habitat} for depth observation. Those pretrained ResNet50 backbones are frozen during training. Sliding is enabled in the evaluation phase.

\begin{figure*}
  \includegraphics[width=\linewidth]{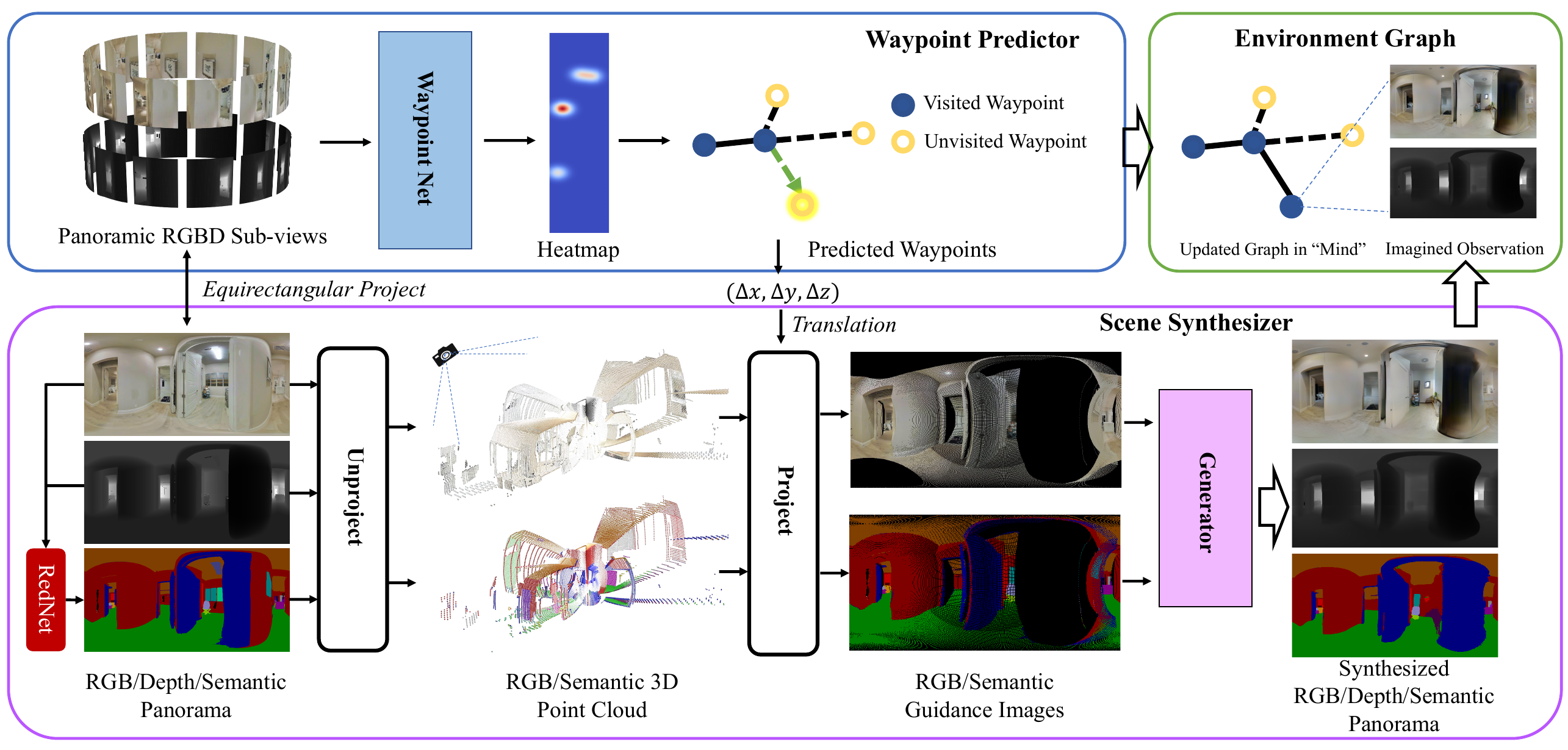}
  \caption{An overview of the world model. The world model enables the agent to imagine the future state of making an action.}
  \label{fig:overview}
\end{figure*}

\noindent\textbf{Details$_{\!}$ of$_{\!}$ World$_{\!}$ Model.} Our$_{\!}$ world$_{\!}$ model$_{\!}$ consists$_{\!}$ mainly$_{\!}$ of a \textsl{Waypoint\_Predictor} (\S\red{3.1}) and a \textsl{Scene\_Synthesizer} (\S\red{3.1}). 
An overview of our world model is illustrated in Figure~\ref{fig:overview}. The \textsl{Waypoint\_Predictor} generates a heatmap based on the panoramic RGBD observation. Each pixel in the heatmap represents the probability of a waypoint's existence in a certain direction and at a certain distance. We use non-maximum suppression (NMS) to limit the number of candidate waypoints to 5. The \textsl{Scene\_Synthesizer} generates a plausible observation for a new waypoint in two steps: 1) The RGBD and semantic map of the initial waypoint, denoted as $v$, are unprojected to 3D space as point cloud. The point cloud is then reprojected to the image plane of the camera at the new waypoint, denoted as $v'$, as guidance images (\ie RGB, depth, and semantic images). 2) The guidance images and the RGB observation at $v$ is fed into a generative network to synthesize RGBD and semantic images for $v'$. It is worth noting that the input of the \textsl{Waypoint\_Predictor} is a panorama represented by 12 separated single-view images, whereas the input of the \textsl{Scene\_Synthesizer} is a single 360$^\circ$ panoramic image. To align the input format, we perform equirectangular projection to convert the subviews to a panoramic image for the \textsl{Scene\_Synthesizer}. Following~\cite{koh2021pathdreamer}, we apply the RedNet~\cite{jiang2018rednet} to predict the initial semantic map based on the RGBD observation. 

\noindent\textbf{Learning of Distance Function.} We utilize the R2R training split for the learning of the distance function $F_d$ (\S\red{3.2}). The learning goal of $F_d$ is to minimize the L2 loss between the predicted distance and the GT (ground-truth) distance of each waypoint in the built EG, \ie environment graph (\S\red{3.1}). Concretely, for each training episode, we progressively build EGs as moving along the GT trajectory. The GT trajectory is the shortest path to the target position following the waypoints predicted by the \textsl{Waypoint\_Predictor} since the topological graphs of environments are not given in VLN-CE. As the EG during mental planning may contain other waypoints besides the GT waypoints, for robust distance prediction, we apply a random strategy to extend the EG during training. Specifically, at each step along the GT trajectory, we extend the EG by adding the next best waypoint and at most 5 other random sampled waypoints accessible from the waypoints in the current graph. The training loss of an episode is formulated as
\begin{equation}\small
  \text{Loss} = \sum_{\mathcal{G}\in\mathfrak{G} }\sum_{v\in\mathcal{V}} \parallel F_d(v,\mathcal{G},X)-d_v\parallel,
\end{equation}
where $\mathfrak{G} $ is the set of all constructed EGs in this episode, $\mathcal{V}$ is the waypoint set of $\mathcal{G}$, $d_v$ is the GT distance of the waypoint $v$, and $X$ is the language instruction of this episode.
As $F_d$ predicts the distances of waypoints based on the synthesized observations during mental planning, we randomly replace GT observations in EG with observations synthesized by \textsl{Scene\_Synthesizer} to make $F_d$ better adapted to inference. For fast convergence, we first train the network using the GT observations for 10 epoches and then finetune the network on EGs where the observations of some waypoints are ranodmly replaced by the synthesized ones for another 10 epoches. We adopt AdamW optimizer~\cite{loshchilovdecoupled} for network training with the learning rate set to 2.5e-5 and set the batch size to 16. We apply max clip gradient normalization to all parameters to stabilize the training.

\noindent\textbf{Mental Planning.} For a comprehensive understanding of the mental planning procedure, we present a python-style pseudo code in Alg.~\ref{alg:dw}. Our \textsc{Dreamwalker} agent is equipped with a world model to ``imagine'' state transitions of taking actions, and a distance function for reward prediction and fast rollout. To perform mental planning for a step of decision making, the agent calls \texttt{search} function with the current EG \texttt{initState}, the discount factor \texttt{gamma} and the exploration constant \texttt{C} as arguments, and gets the best action as the next action to make after rounds of search. The factor \texttt{gamma} used in experiments is 0.98 and \texttt{C} is 1.0.

\begin{algorithm*}
  \caption{A python-style pseudo code of our \textsc{Dreamwalker}.}
      \label{alg:dw}
      \vspace{5pt}
  \includegraphics{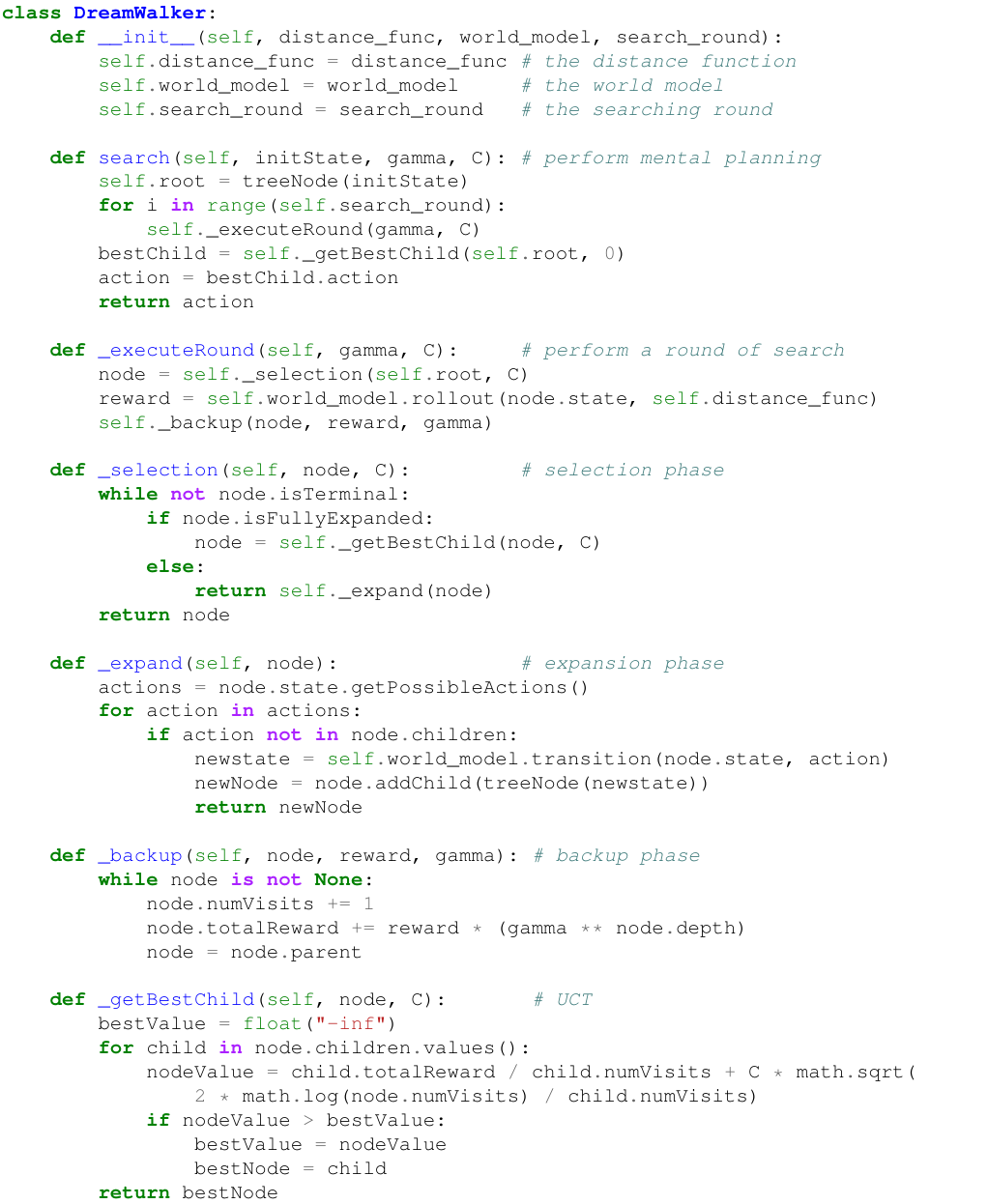}
  \vspace{15pt}
\end{algorithm*}

\section{Qualitative Results} 
\label{sec:qualitative}

\begin{figure*}
  \centering
  \includegraphics[width=0.93\linewidth]{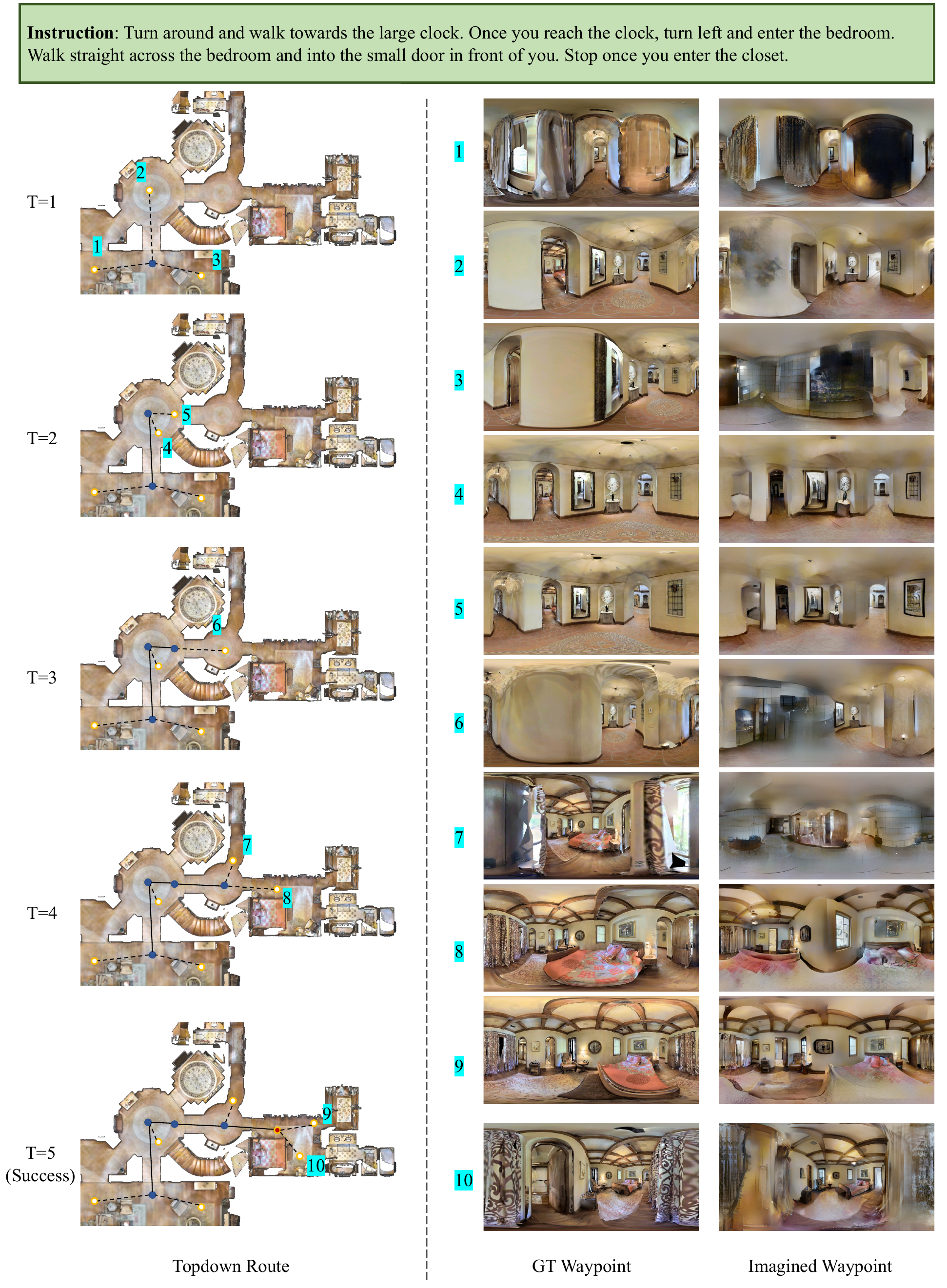}
  \caption{\textsc{Dreamwalker} navigates by imagining the future states of executing an action.}
  \label{fig:case}
\end{figure*}

In this section, we present some qualitative results of navigation process on the \texttt{val\_unseen} split of R2R dataset. As shown in Figure~\ref{fig:case}, our \textsc{Dreamwalker} can perform meaningful and high-quality imagination of future states (\eg waypoint \#2, \#5, and \#8) based on the current observation to facilitate robust decision-making. It is worth noting that the coordinate of the predicted waypoint could be different from the actual reached waypoint as the agent may collide with some obstacles during moving.

\begin{figure*}
  \centering
  \includegraphics[width=0.90\linewidth]{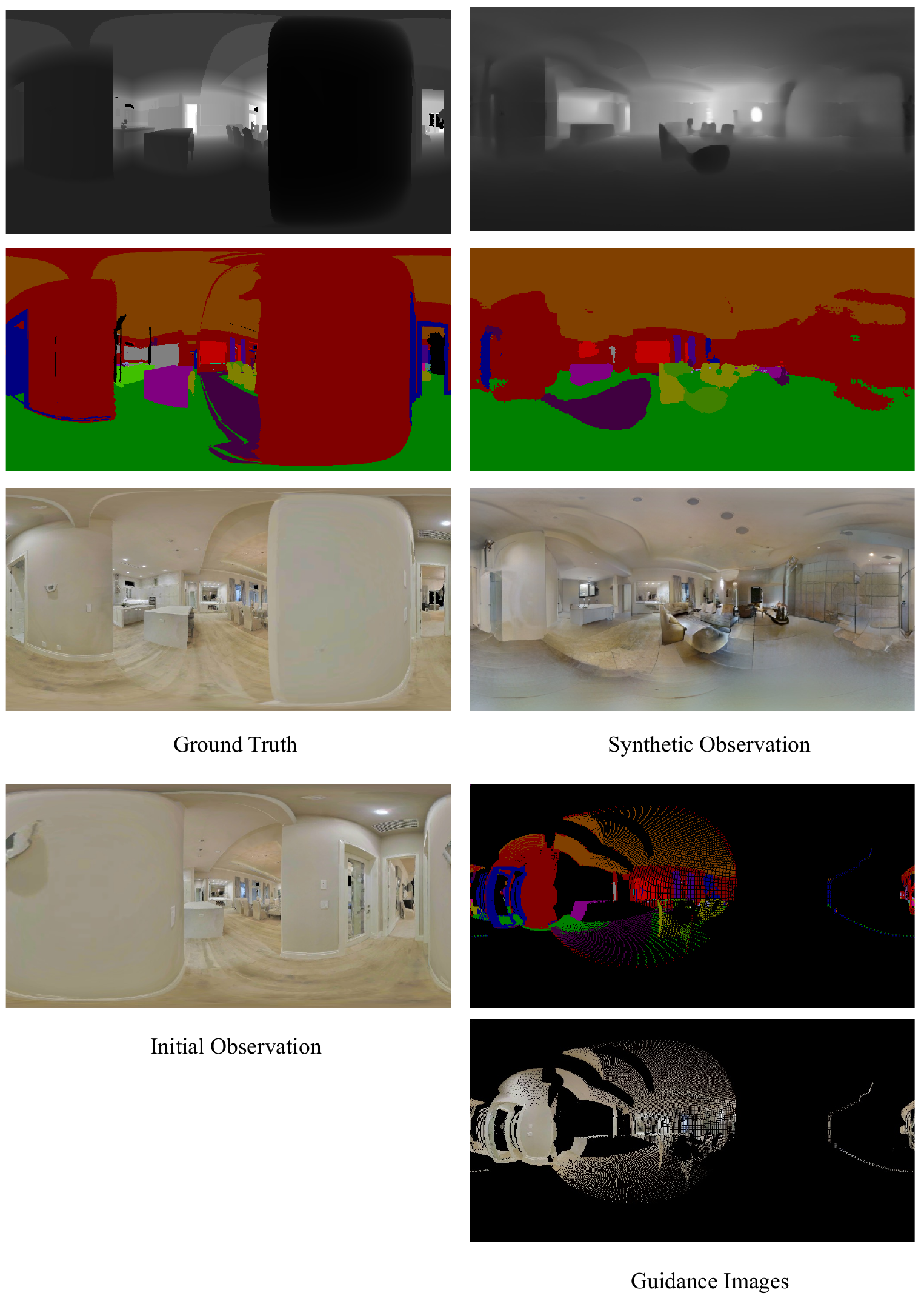}
  \caption{A threading up case.}
  \label{fig:flaw1}
\end{figure*}

\begin{figure*}
  \centering
  \includegraphics[width=0.90\linewidth]{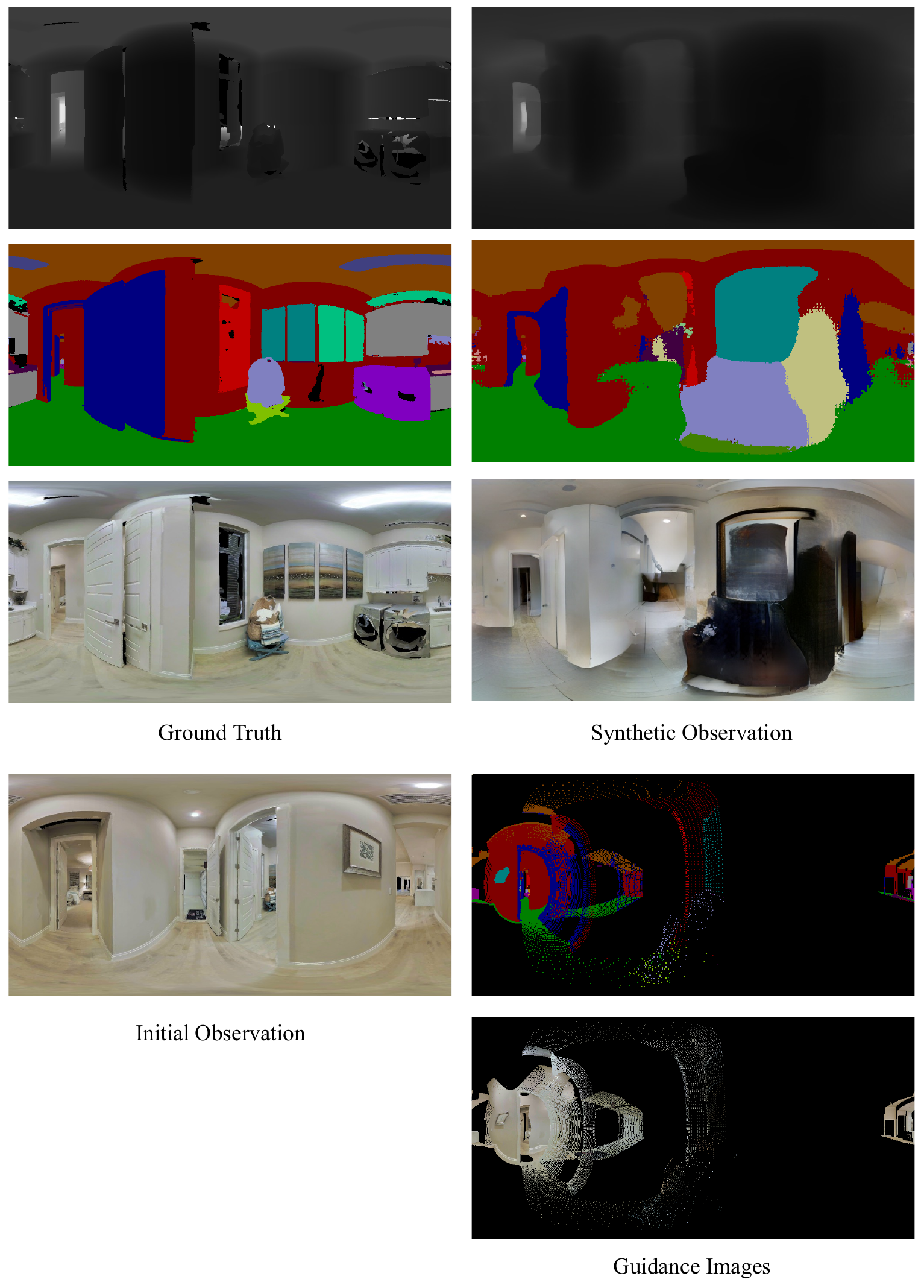}
  \caption{An occlusion case.}
  \label{fig:flaw2}
\end{figure*}

\begin{figure*}
  \centering
  \includegraphics[width=0.90\linewidth]{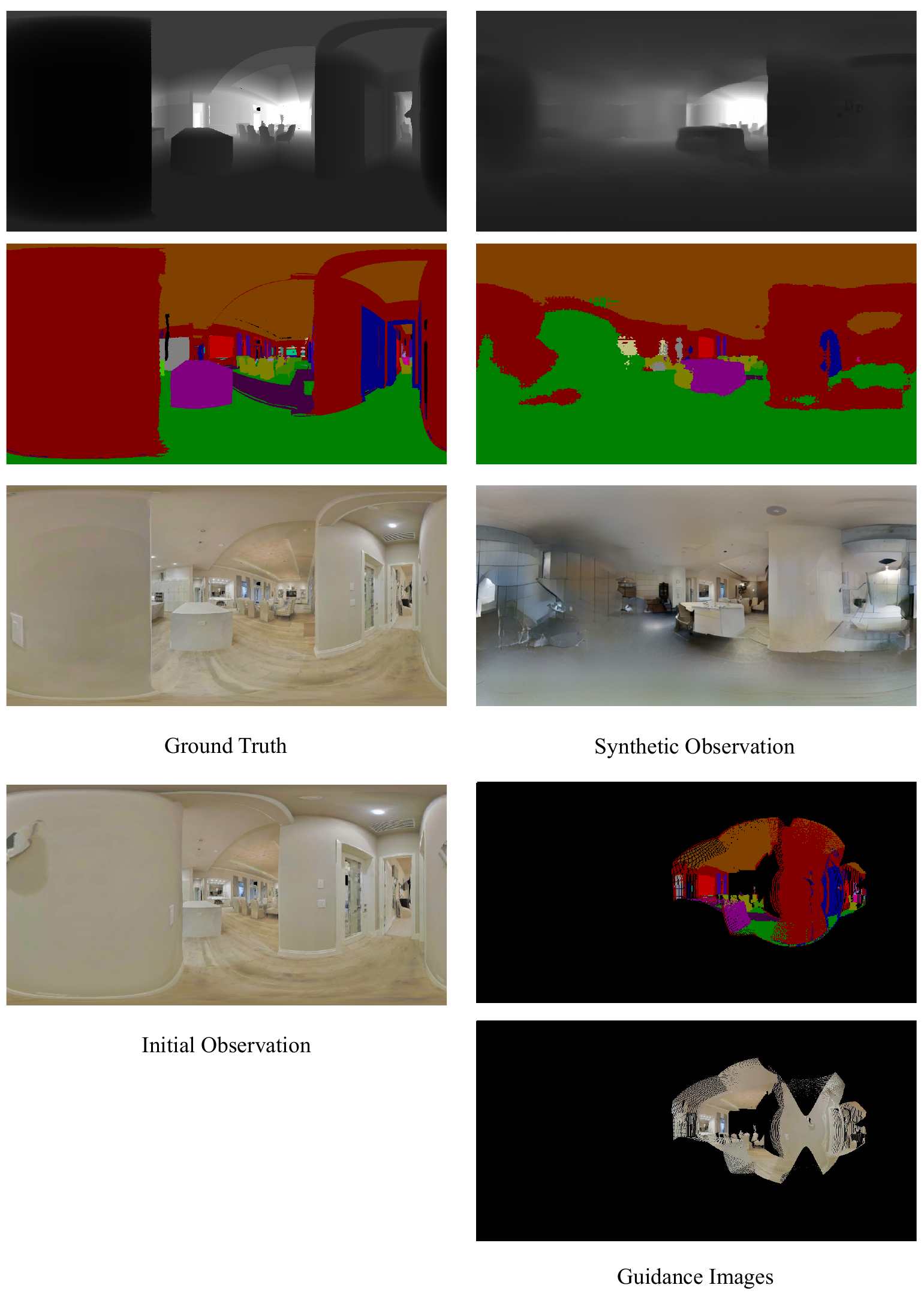}
  \caption{A case of predicing a distant waypoint.}
  \label{fig:flaw3}
\end{figure*}

We also observe some flaws of synthesized observations. In particular, the synthesized observation is blurry when the projection of the point cloud is sparse. The reason for this issue is three-fold:
\begin{itemize}
  \setlength{\itemsep}{0pt}
  \setlength{\parsep}{0pt}
  \setlength{\parskip}{0pt}
  \item The new waypoint is greatly occluded from the initial location. It means that the details of the room is not observable from the opposite perspective, which leads to sparse guidance images. As shown in Figure~\ref{fig:flaw2}, the details of the room is greatly occluded by the wall.
  \item The new waypoint is far away from the initial location. Due to the perspective principle, the point cloud is projected into a small range of pixels in the guidance images. A case is illustrated in Figure~\ref{fig:flaw3}.
  \item The predicted new waypoint is threaded up the mesh of the environment. In this situation, the predicted waypoint is detrimental to navigation. As shown in Figure~\ref{fig:flaw1}, the predicted waypoint locates inside the wall.
\end{itemize}

Several workarounds may help alleviate the aforementioned issues. For instance, to prevent occlusions and threading up, a collision detection and avoidance mechanism could be developed to identify if a predicted waypoint would result in a collision with the environment's mesh. Should a collision be detected, the algorithm can recompute an alternative waypoint or adjust the current one to circumvent the obstacle. Moreover, further study the optimal maximum distance can help mitigate the sparse projection of guidance images caused by long distance point cloud translations. We consider addressing these issues as part of our future work.



\section{Discussions}
\label{sec:discuss}

\noindent\textbf{Limitations.} Our work represents an initial effort that investigates world models in the context of challenging vision-language embodied tasks. As such, there are many aspects of this framework that warrant further study. We identify the limitations of our work in terms of world model, mental planning, and real-world applicability. At present, the \textsl{Waypoint\_Predictor} and the \textsl{Scene\_Synthesizer} are trained on the training split of Matterport3D~\cite{Matterport3D} dataset, which comprises only 61 scenes. The limited training data restricts the quality and diversity of synthesized observations, particularly in terms of layout and appearance. Adding constraints such as interpenetration detection could improve these waypoint predictions. Furthermore, the quality of the synthesized views can be enhanced by replacing the scene synthesizer with diffusion-based methods~\cite{hoellein2023text2room}.

The world model based mental planning may encounter computational challenges when scaling up to larger, more complex environments. The increased computational demands for simulating and evaluating potential plans could probably slow down the agent's decision-making process and decrease its overall efficiency. This high computational burden can be mitigated by maintaining a sliding buffer of navigation history.

Currently, the agent navigates within static virtual environments. However, in real-world scenarios, its performance may be affected by dynamic elements in the environment, such as moving objects or changing conditions (\eg, lighting or crowdedness). Developing strategies to address these dynamic changes is crucial for the agent's effectiveness in real-world settings, and this will be considered in future work.

\noindent\textbf{Social Impact.} 
\textsc{Dreamwalker}'s ability to make strategic planning through mental experiments can lead to more intelligent and interpretable decision-making, which can be beneficial when translating the agent to real-world applications. The agent's capacity to simulate future scenarios and make its decision-making process more transparent can build trust in the technology, making it more acceptable and useful to users. However, the current use of a simulated platform for testing and development may result in discrepancies between the simulation and real-world environments. This can limit the direct applicability of the agent's performance in real-world scenarios, potentially requiring further adaptation and fine-tuning. Additionally, as AI agents like \textsc{Dreamwalker} continue to advance, it is crucial to consider the importance of accessibility and inclusivity in the design and implementation of these systems. If the technology is not developed with diverse user needs in mind, it may unintentionally exclude certain groups of people, such as those with disabilities or those from non-English speaking backgrounds. Ensuring that the agent can understand and process various languages, dialects, and cultural nuances, as well as catering to the needs of individuals with differing abilities, is essential for promoting equitable access to the benefits of embodied robots.

{\small
\bibliographystyle{ieee_fullname}
\bibliography{egbib}
}

\end{document}